\documentclass[journal]{IEEEtran}
\usepackage{cite}
\usepackage{amsmath,amssymb,amsfonts}
\usepackage{algorithmic}
\usepackage{graphicx}
\usepackage{textcomp}
\usepackage{multirow}
\usepackage{multicol}
\usepackage{url}
\usepackage{hyperref}

\usepackage[dvipsnames]{xcolor}

\newcommand{\beginsupplement}{%
        \setcounter{table}{0}
        \renewcommand{\thetable}{S\arabic{table}}%
        \setcounter{figure}{0}
        \renewcommand{\thefigure}{S\arabic{figure}}%
     }

\def\BibTeX{{\rm B\kern-.05em{\sc i\kern-.025em b}\kern-.08em
    T\kern-.1667em\lower.7ex\hbox{E}\kern-.125emX}}
\begin{document}
\title{Toward Interpretable Sleep Stage Classification Using Cross-Modal Transformers}
\author{Jathurshan Pradeepkumar $^{\star}$ , Mithunjha Anandakumar $^{\star}$, Vinith Kugathasan, Dhinesh Suntharalingham,\\ Simon L. Kappel, Anjula C. De Silva and Chamira U. S. Edussooriya  
\thanks{$^{\star}$ These authors contributed equally to the work.}
\thanks{ Jathurshan Pradeepkumar and Mithunjha Anandakumar were with the Department of Electronic and Telecommunication
Engineering, University of Moratuwa, Sri Lanka and currently affiliated to Faculty of Arts and Sciences at Harvard University. Vinith Kugathasan, Dhinesh Suntharalingham, Anjula C. De Silva and Chamira U. S. Edussooriya are with the Department of Electronic and Telecommunication
Engineering, University of Moratuwa, Sri Lanka. Chamira U. S. Edussooriya is also with the  Department of Electrical and Computer Engineering, Florida International University, Miami, FL, USA. Simon L. Kappel is with Department of Electrical and Computer Engineering, Aarhus University, DK-8200 Aarhus, Denmark.\\
Corresponding Author: Chamira U. S. Edussooriya (chamira@uom.lk) \\ 
This work was supported in part by the Senate Research Committee, University of Moratuwa.\\
© 2024 IEEE.  Personal use of this material is permitted.  Permission from IEEE must be obtained for all other uses, in any current or future media, including reprinting/republishing this material for advertising or promotional purposes, creating new collective works, for resale or redistribution to servers or lists, or reuse of any copyrighted component of this work in other works.”}}

\maketitle

\begin{abstract}
Accurate sleep stage classification is significant for sleep health assessment. In recent years, several machine-learning based sleep staging algorithms have been developed , and  in particular, deep-learning based algorithms have achieved performance on par with human annotation. Despite improved performance, a limitation of most deep-learning based algorithms is their black-box behavior, which have limited their use in clinical settings. Here, we propose  a \emph{cross-modal transformer}, which is a transformer-based method for sleep stage classification.  The proposed cross-modal transformer consists of a cross-modal transformer encoder architecture along with a multi-scale one-dimensional convolutional neural network for automatic representation learning. The performance of our method is on-par with the state-of-the-art methods and eliminates the black-box behavior of deep-learning models by utilizing the interpretability aspect of the attention modules.  Furthermore, our method provides considerable reductions in the number of parameters and training time compared to the state-of-the-art methods. Our code is available at \url{https://github.com/Jathurshan0330/Cross-Modal-Transformer}. A demo of our work can be found at \url{https://bit.ly/Cross_modal_transformer_demo}.
\end{abstract}

\section{Introduction}
\label{sec:Introduction}



Accurate sleep stage classification plays a crucial role in sleep medicine and human health. In general, sleep experts use polysomnography (PSG) recordings to diagnose sleep related disorders. PSG mainly comprises of electroencephalogram (EEG) and electrooculogram (EOG). The PSG recordings are typically segmented into $30$ s epochs and manually annotated by sleep experts based on guidelines such as Rechtschaffen and Kales (R\&K)~\cite{rechtschaffen1968manual} or American Academy of Sleep Medicine (AASM)~\cite{Iber2007}. The manual annotation process is tedious, prone to human errors, labour intensive and time consuming. To overcome these drawbacks, multiple studies have proposed automatic sleep stage classification algorithms as the alternative. 


Recent works that employ deep learning based algorithms achieved impressive results in sleep stage classification~\cite{loh2020} than the conventional machine learning algorithms such as support vector machines (SVMs)~ \cite{RAHMAN2018211}, k-nearest neighbors (KNNs),  decision trees~\cite{HASSAN2016248} and random forest classifiers~\cite{RAHMAN2018211}. These include convolutional neural networks (CNNs)~\cite{b5,b6,CNN1,CNN2}, recurrent neural networks (RNNs)~\cite{RNN1,RNN2,BHI_RNN}, deep belief networks (DBNs)~\cite{DBN1}, autoencoders~\cite{AE1} and hybrid architectures, such as CNN with RNN \cite{Hyb1,supratak2017} and deep neural networks with RNN~\cite{Hyb3}. Furthermore, different learning techniques such as transfer learning \cite{HUYPHAN}, multi-view learning \cite{xsleepnet}, meta learning~\cite{meta_2021}, knowledge distillation~\cite{KD_2022} and model personalization have improved the performance of automatic sleep staging. With these advances, automatic sleep staging have been able to achieve a performance on par with manual annotation.

Despite the improved performance, a significant limitation lies in their black-box behavior. A general concern when it comes to application of artificial intelligence in healthcare and medicine is the underlying mechanism of deep-learning algorithms \cite{lahav2018interpretable}. This is a major drawback of existing deep learning based sleep staging algorithms that keeps them from being adopted towards clinical settings.
To address this issue, we focus on developing \emph{interpretable} sleep staging algorithms in our work An interpretable model has the ability to provide explanations for the decisions made concerning specific inputs, thereby offering a key solution to this problem.


Transformers~\cite{vaswani2017} has become \textit{de facto} in natural language processing (NLP) tasks. Following the seminal work \cite{vaswani2017}, there have been exceptional works on transformers such as BERT \cite{Bert}, vision transformers (ViTs) \cite{ViT}, which have improved the state-of-the-art in both NLP and computer vision domains. Transformers have introduced interpretability aspects in these domains without compromising high performance. To leverage the capability of attention mechanism in transformers for interpretability, we employ transformer based architectures on physiological signals in our method, specifically for automatic classification of sleep stages to achieve high prediction performance together with interpretability. To the best of our knowledge, SleepTransformer~\cite{Trans1} and SleepViTransformer~\cite{peng2023sleepvitransformer} are the only works which has explored transformers for sleep stage classification. Transformer encoders used in SleepTransformer~\cite{Trans1} are inspired from the seminal transformer architecture proposed in~\cite{vaswani2017}, and on top of the transformer encoder they utilize an additional attention layer to achieve a compact representation for each epoch of the PSG signals. We hypothesized that a better compact representation can be learned with less overhead by employing an additional learnable vector named $[CLS]$ vector similar to BERT \cite{Bert} and ViTs \cite{ViT}. The vector representation learned by the transformer encoder corresponding to the $CLS$ vector can be utilized as the compact vector representation for each epoch. This learned vector representation aggregates all the sequence information in a PSG epoch. Another drawback of SleepTransformer~\cite{Trans1} is that they only utilize EEG signals for sleep staging, whereas in our method we explore the capability of transformers in learning from multiple signals. The SleepTransformer~\cite{Trans1} learns only intra and inter epoch relationships in an EEG signal whereas in our method, we explore cross-modal relationships between EEG and EOG signals along with intra-modal and inter epoch relationships, by modifying our transformer encoder architecture.

\begin{figure}[!t]
 \centering
    \includegraphics[width = 1 \linewidth]{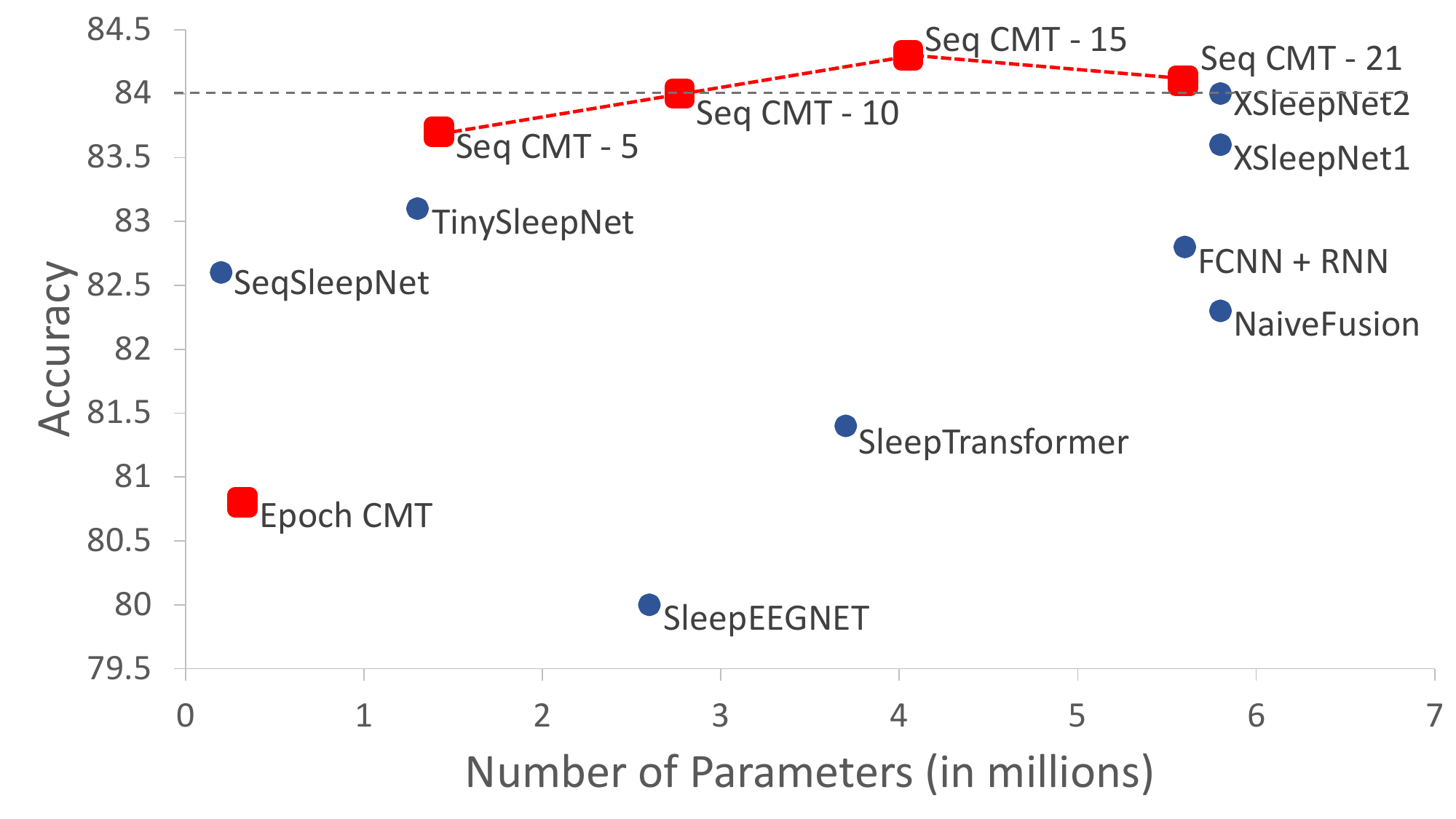}
    \vspace{-0.6cm}
\caption{Performance of our cross-modal transformers (in red squares) and other previously reported works (in blue circles) on sleep-EDF-expanded 2018 dataset. Our sequence cross-modal transformer achieves on-par performance with the state-of-the-art, with fourfold reduction in parameters. Here, Seq and CMT refer to Sequence and cross-modal transformer.}
\label{fig:plot_comparison}
\vspace{-0.5cm}
\end{figure}

To overcome the limitations of past works and to improve interpretability of deep learning based sleep staging algorithms, we propose \emph{cross-modal transformers}. Cross-modal transformers consists of a cross-modal transformer encoder architecture along with a multi-scale one-dimensional convolutional neural network (1D CNN) for automatic representation learning. The major contributions of our work presented in this paper are summarized below:
\begin{itemize}
    \item We propose two cross-modal transformers: 1) Epoch cross-modal transformer and 2) Sequence cross-modal transformer to solve the problem of sleep stage classification under two classification schemes which are one-to-one and many-to-many classification.
    \item Our method employs a two-stage process: initially, it learns representations directly from the  raw signals of each modality, which then followed by our cross-modal transformer architecture. We utilize existing multi-scale 1D-CNN architecture and adapted towards our specific task in the initial stage to learn an optimal feature representation by considering both local and global features.
    \item A cross-modal transformer encoder architecture to learn both intra-modal temporal attention, i.e., attention between time steps within a feature representation of a modality, and cross-modal attention, i.e., attention between each modalities. The sequence cross-modal transformer consists of an additional block to learn inter-epoch attention, i.e., attention between adjacent epochs.
    \item A simple yet an effective method based on attention mechanisms to enable interpretation of the model's predictions. The proposed method is capable of learning and interpreting: 1) intra-modal relationships, 2) cross-modal relationships and 3) inter-epoch relationships. 
     \item The performance of our sequence cross-modal transformers is on-par with the state-of-the-art methods with a comparable reduction in the number of parameters as can be seen in Fig.~\ref{fig:plot_comparison}.
     

    \item Our method requires lesser time to train and has a smaller model footprint in terms of parameters compared to the state-of-the-art methods.
\end{itemize}

\vspace{-0.5em}

\section{Related Work}
\label{sec:related}

\subsection{Transformers}
Since the introduction by \textit{Vaswani et al.} \cite{vaswani2017}, \emph{Transformers}  and its variants have been successful in NLP and computer vision tasks in terms of performance and powerful representation learning. Currently, Transformers have achieved the state-of-the-art performance in several NLP tasks. Pre-training have been leveraged in Transformers to achieve better performance in NLP tasks, where BERT \cite{Bert} utilized self-supervised pretraining strategies and fine-tuning towards a supervised downstream task, whereas GPT \cite{radford2018} focused on language modeling in pre-training. Although CNNs have dominated the field of computer vision, there have been several efforts to explore the benefits of Transformers in computer vision domain. Initially, self-attention was introduced along with CNNs for computer vision tasks \cite{yuan2018ocnet}. Then the focus was on completely replacing convolutions in the architecture. ViTs \cite{ViT} were able to achieve this feat, by directly applying them on images with very few modifications. Furthermore, self-supervised pre-training \cite{DINO} was leveraged to improve the performance, which emerged several properties of the ViTs. Likewise, Transformers have been explored in several medical applications \cite{application} such as medical image segmentation, detection and registration, whereas SleepTransformer \cite{Trans1} and SleepViTransformer \cite{peng2023sleepvitransformer} are the only works in the domain of sleep staging. SleepTransformer was able to achieve a competitive accuracy of $81.4\%$ on sleep-EDF-expanded 2018  without any pre-training.

\subsection{Deep Learning Based Sleep Stage Classification}

Recent work that employ deep-learning based algorithms achieved impressive results in sleep stage classification~\cite{loh2020} than conventional machine learning algorithms. The conventional machine learning algorithms generally use the handcrafted features as the input \cite{HASSAN2016248,RAHMAN2018211, HASSAN20161}. Furthermore, this approach may not generalize to a large population due to the heterogeneity of subjects and recording devices~\cite{supratak2017}. 

The existing automatic sleep stage algorithms can be classified into two categories based on the input to the network: raw signals and time-frequency maps. A time-frequency map requires prior knowledge of the dataset as well as signal processing, as it heavily relies on the preprocessing steps. 
In contrast, deep-learning based algorithms are capable of performing automatic feature extraction, thus the drawbacks of handcrafted features are eliminated.

Several previous works have focused on utilizing the single channel EEG recordings for sleep stage classification. However, EOG is capable of detecting eye movements, which is a fundamental indicator for differentiating rapid eye movement (REM) and non-rapid eye movement (NREM) stages~\cite{sun2019}. Therefore, the information in EOG channels, i.e., multi-modal scheme can be exploited to improve the performance of sleep stage classification~\cite{RAHMAN2018211},\cite{mikkelsen2021}. 

\textit{Phan et al.} \cite{xsleepnet}, have used both raw signals and time-frequency representations of the signals as the inputs to a sequence-to-sequence network. The overall accuracy achieved was $84\%$  with only single channel EEG, as well as with both single channel EEG and EOG. XSleepNet is the current state-of-the-art in automatic sleep staging. The inclusion of EOG signal is insignificant in terms of the performance of the XSleepNet model. 
Supratak et al.~\cite{supratak2017} have proposed CNN + LSTM to capture time invariant features and transition characteristics among sleep stages from raw single channel EEG. Mousavi et al.~\cite{mousavi2019} have proposed SleepEEGNet, composed of CNN to extract features and sequence-to-sequence model to capture long short-term context dependencies between epochs of raw single channel EEG. The sequence-to-sequence model composed of bidirectional RNN and attention mechanisms. Even though RNN and LSTM based approaches yields a competitive performance with state-of-the-art methods, the computational cost of LSTMs and RNNs should be taken into consideration. RNNs have limitations due to their recurrent nature and high complexity, thus difficult to train them in parallel. Eldele et al.~\cite{eldele2021} have proposed AttnSleep which consists of a multi-resolution CNN with adaptive feature re-calibration and temporal context encoder utilizing multi-head self attention mechanism to capture the temporal dependencies from single channel EEG. AttnSleep achieved accuracy of $82.9\%$, which is higher than most of the existing algorithms. SeqSleepNet, proposed by Phan et al.\cite{phan2019seqsleepnet}, is a sequence-to-sequence hierarchical RNN classification model trained on multi-channel time-frequency maps as input. SeqSleepNet uses attention based bidirectional RNN for short-term sequence modeling and bidirectional RNN for long-term sequence modeling. SeqSleepNet achieved accuracy of $82.6\%$ with single channel  EEG and 83.8\% with both single channel EEG and EOG channels demonstrating the significance of including EOG to improve the accuracy. TinySleepNet proposed by Supratak et al.~\cite{supratak2020tinysleepnet} is an efficient model for sleep stage classification based on raw single channel EEG, which achieved competitive performance with $83.1\%$ accuracy. TinySleepNet is an improved version of DeepSleepNet model with improved efficiency, i.e., less number of parameters, thus reducing the computational resource requirement.



\begin{figure}[t]
    \centering
    \includegraphics[width = 0.8 \linewidth]{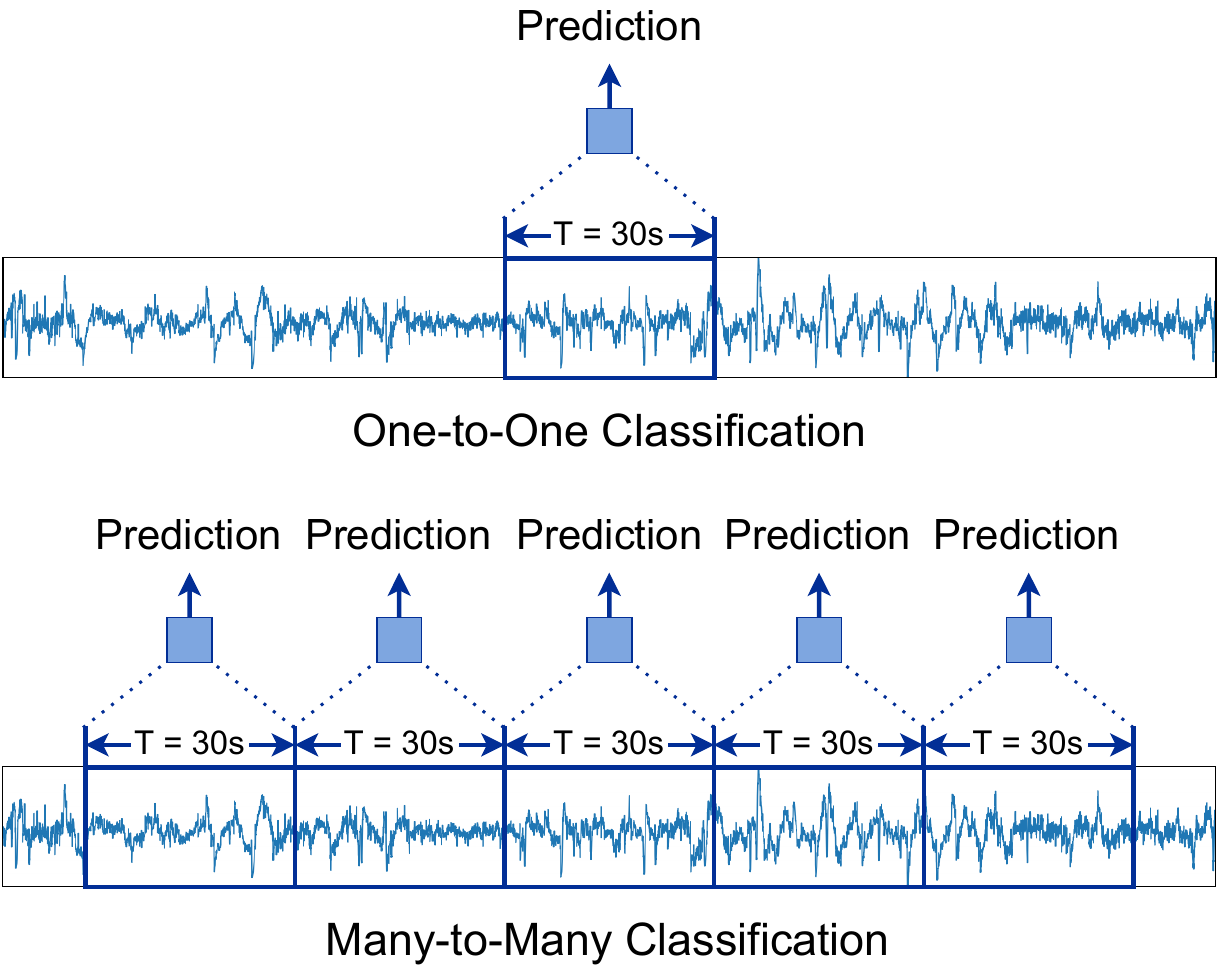}
    \caption{The two classification schemes\cite{phan2019seqsleepnet} used in the domain of sleep staging and in our experiments. In one-to-one classification, the sleep stage of an individual PSG epoch is predicted, whereas in many-to-many classification the sleep stages of multiple epochs are predicted simultaneously.}
    \label{fig:classi}
    \vspace{-0.6cm}
\end{figure}

\begin{figure*}[!t]
 \centering
    \includegraphics[width = 1 \linewidth]{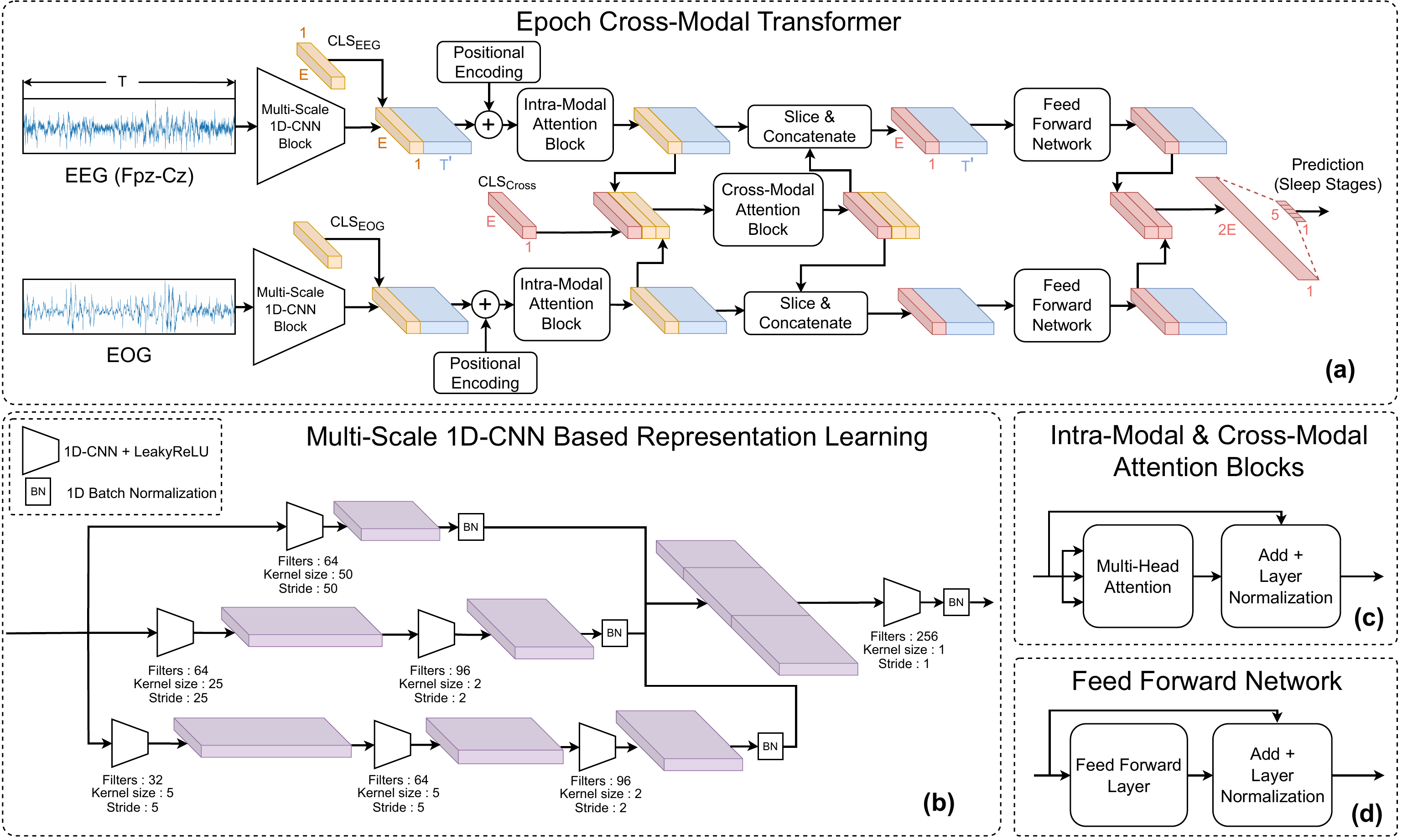}
    \caption{The architecture of the epoch cross-modal transformer consisting of multi-scale 1D-CNN blocks, intra-modal attention blocks, cross-modal attention block and feed forward networks. (a) shows the overall architecture with two signals as input. (b) visualizes multi-scale 1D-CNN blocks, which consists of three pathways to learn both local and global features. (c) and (d) shows the architectures of the attention blocks and feed forward networks. Here $CLS_{EEG}, CLS_{EOG}$ and $CLS_{Cross}$ are the $CLS$ vectors initiated to learn the aggregated representation of intra-modal relationships of EEG and EOG modalities and cross-modal relationship between EEG and EOG.}
\label{fig:EpochCMT}
\vspace{0.5cm}
\end{figure*}

\begin{figure*}[!t]
 \centering
    \includegraphics[width = 1 \linewidth]{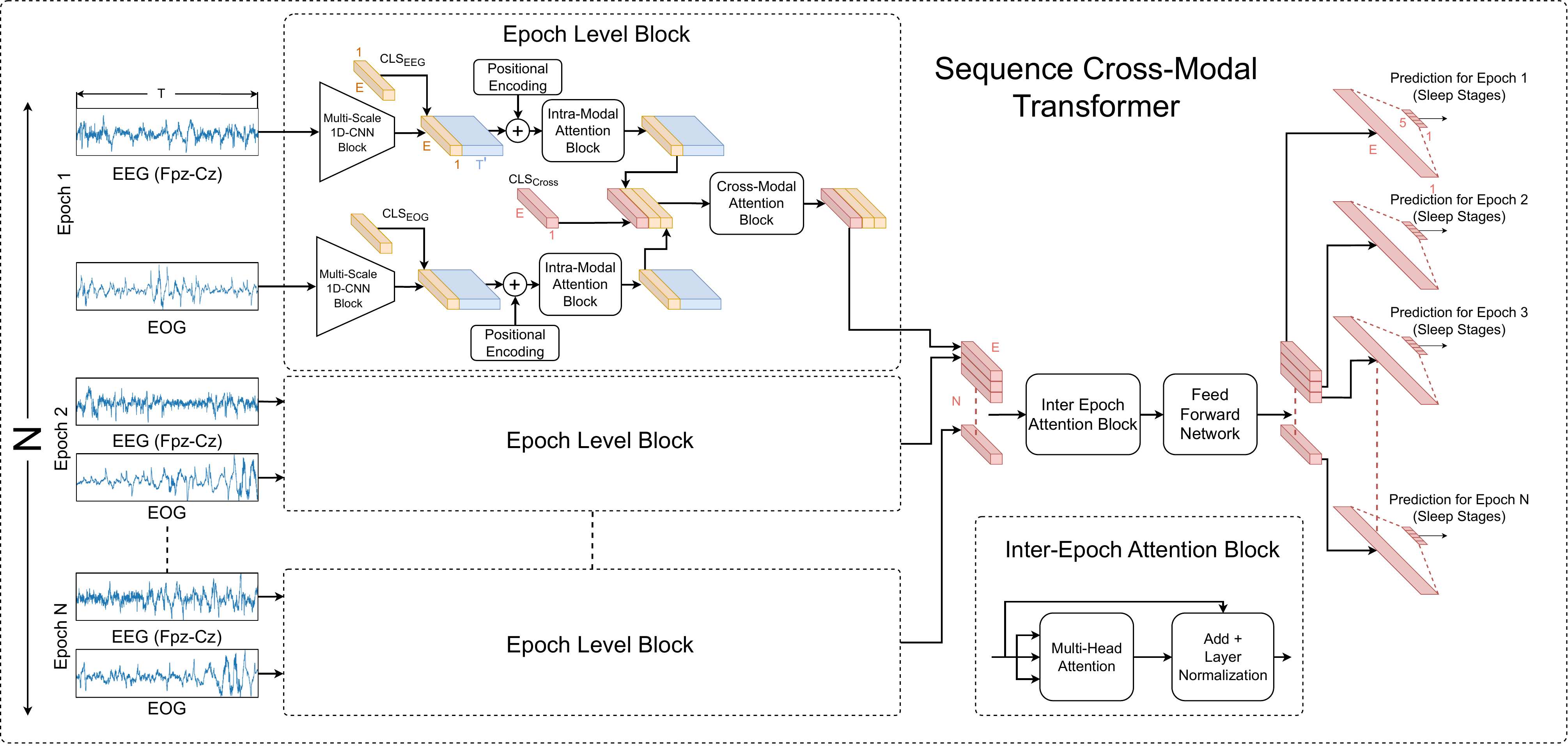}
\caption{The architecture of sequence cross-modal transformer, which is an extension of epoch cross-modal transformer. The sequence cross-modal transformer consists of multiple epoch level blocks to learn the epoch level representation and an additional block to learn inter epoch relationships.}
\label{fig:SeqCMT}
\vspace{+1cm}
\end{figure*}

\vspace{-0.1cm}
\section{Methodology}
\label{sec:Methodology}
In this section, our proposed cross-modal transformers are presented. First, we formulate the problem definition for one-to-one and many-to-many sleep stage classification tasks. Then we introduce our cross-modal transformers: the epoch cross-modal transformer for one-to-one classification and the sequence cross-modal transformer for many-to-many classification. We further explain the multi-scale 1D CNN based representation learning and cross-modal transformer encoder under the epoch cross-modal transformer.


\vspace{-0.3cm}
\subsection{Problem Definition}
We address the problem of classifying the sleep stage of a $30$~s epoch of PSG signals which were acquired in the experiment described in section~\ref{sec:Dataset}. Our training set with the size of $N$, consists of labeled $30$~s PSG epochs $\{x_{i},y_{i}\}_{n=1}^{N}$, where $\{x_{i},y_{i}\}$ $\in$ $X \times Y$. Here $X$ $\in$ $\mathbb{R}^{T \times C}$ denotes the input space of recorded PSG signals, where $T$ represents the time steps in an epoch and $C$ $\in$ \{EEG (Fpz-Cz), EOG\} represents the modalities in the recorded PSG signals. $Y$ $\in$  \{WAKE, N1, N2, N3, REM\} represents the output space of sleep stages. Our goal is to learn a function $f_\theta : X \xrightarrow{} Y$ by minimizing the error $\mathbb{E}_{(x,y)}(\mathcal{L}(f_{\theta}(x),y))$ on the given training dataset. Here, $\mathcal{L}$ denotes the loss function.

We solve the aforementioned problem using cross-modal transformers under the one-to-one and many-to-many classification schemes{\cite{phan2019seqsleepnet} illustrated in  Fig.~\ref{fig:classi}. In the one-to-one classification, we consider a single PSG epoch to predict the corresponding sleep stage \cite{RNN1}. In many-to-many classification, we consider a sequence of PSG epochs and predict their corresponding sleep stages at once \cite{RNN1}.

\subsection{Epoch Cross-Modal Transformer}
In this subsection, we focus on solving sleep stage classification with the one-to-one scheme using a epoch cross-modal transformer. The proposed epoch cross-modal transformer, shown in Fig.~\ref{fig:EpochCMT}, consists of two main blocks: 1) multi-scale 1D-CNN for automatic representation learning and 2) a cross-modal transformer encoder architecture to learn both intra-modal temporal attention and cross-modal attention. The epoch transformer initially learns two separate feature representations from the input EEG and EOG signals in an epoch using 1D-CNN. Here, features are learned and extracted using non-overlapping windows from the signals. Then, the cross-modal transformer encoder learns a representation by considering intra-modal temporal attention and cross-modal attention, which is then fed into a linear layer for classification. The two main blocks of the epoch transformer are further elaborated in proceeding subsections. The key advantage of epoch cross-modal transformer is that it smaller in size and enables faster training and tuning of the hyper-parameters on a resource constraint environment. The same model can easily be scaled into a sequence cross-modal transformers efficiently.

\subsubsection{Multi-Scale 1D-CNN for Representation Learning}

Inspired by vision transformers using image patches as sequential data\cite{ViT}, we employ multi-scale 1D-CNN as shown in Fig.~\ref{fig:EpochCMT} to learn the feature representation of non-overlapping windows with the size of $0.5$ s from the input $30$s epoch of raw PSG signals. Here, the 1D signals $X_c$ $\in$ $\mathbb{R}^{T\times1}$ will be mapped into a feature space of $X^{'}_{c}$ $\in$ $\mathbb{R}^{(T/(0.5\times f_s))\times E}$, where $c$ $\in$ $C$, $f_s$ is the sampling frequency and $E$ is the embedding size. Let $T/(0.5\times f_s)$  be $T^{'}$ for convenience. The features are extracted from non-overlapping windows instead of overlapping windows to improve interpretability, such that the extracted feature vectors can be fed into the cross-modal encoder as sequential data to learn the attentions between all windows. 

We hypothesize that the global and local features in a window of raw signals will contribute towards the classification of the sleep stages. In order to extract both global and local features in a window, we employed multi-scale 1D-CNN, where the raw input signal goes through three parallel paths as shown in Fig.~\ref{fig:EpochCMT}(b): 1) One 1D-CNN with kernel size of $50$, 2) Two 1D-CNNs with kernel sizes of $25$ and $2$ respectively, and 3) Three 1D-CNNs with kernel sizes of $5$, $5$ and $2$. Each 1D convolution layer will be followed by a LeakyReLU activation. The features extracted at different scales will then be normalized using batch normalization. Finally, the extracted feature representations will be concatenated along the embedding dimension and will undergo an additional 1D-CNN with kernel size of $1$, followed by LeakyReLU activation and batch normalization.

\subsubsection{Cross-Modal Transformer Encoder and Classification}
We propose a cross-modal transformer encoder architecture to learn a powerful feature representation by attending intra-modal temporal information and cross-modal relationships. As illustrated in Fig.~\ref{fig:EpochCMT}(a), the cross-modal transformer encoder consists of two main blocks: 1) attention and 2) position wise fully-connected feed-forward network blocks. 

Initially, for each modality $C$, a learnable $CLS_c$ vector $\in$ $\mathbb{R}^{(1\times E)}$, similar to the one proposed in BERT \cite{Bert}, is randomly initiated and concatenated with the output of the multi-scale 1D-CNN block along the time axis. Similar to the seminal work \cite{vaswani2017}, positional encodings are added to the concatenated vector and fed to the intra-modal attention block to learn the relationships between all the time steps in the modality. We extract only the vector representation corresponding to the $CLS_c$ vector from the output from each modality, as it aggregates all the information of the intra-modal temporal information \cite{Bert}. A new learnable $CLS$ vector named $CLS_{cross}$ is randomly initialized and concatenated along with the $CLS_c$ vector representation extracted from each modality. Cross-modal attention block is employed on the aforementioned vector to learn the relationships between the modalities, which get aggregated in the corresponding representation of $CLS_{cross}$. The vector representation corresponding to the $CLS_{cross}$ is extracted and $CLS_c$ representation of each modality is replaced by this vector. Then the concatenated vectors are passed through the feed-forward network. Finally, the vector representations corresponding to $CLS_{cross}$ in each modality are extracted, flattened and passed through a single linear classifier with $5$ neurons for classification. 

The operations of the intra-modal attention block, cross-modal attention block and feed-forward network are similar to the seminal work by \cite{vaswani2017}. The attention blocks consist of multi-head attention followed by residual addition and layer normalization. The feed-forward network consists of a feed-forward layer followed by residual addition and layer normalization. An additional linear layer is sufficient for classification of the sleep stages from the extracted representation, which shows the power of the learned representation.

\vspace{-0.3cm}
\subsection{Sequence Cross-Modal Transformer}
 In order to solve sleep stage classification with the many-to-many scheme we employ sequence cross-modal transformer, which is an extension of the epoch cross-modal transformer. The proposed architecture shown in Fig.~\ref{fig:SeqCMT} consists of multiple epoch level block for each $30$s epoch of raw PSG signals in a non-overlapping sequence (Fig.~\ref{fig:classi}). The epoch level block is built using multi-scale 1D-CNN block and intra-modal attention block for each modality and a cross-modal attention block. The corresponding vector representation of $CLS_{cross}$ vector of each epoch block is extracted and concatenated. Inter-epoch attention block followed by a feed-forward network is employed on the concatenated vector to learn the relationships between the epochs. Finally, we extract and flatten output representation corresponding to each epoch separately and employ a linear classifier to predict their sleep stages. During inference, we let the model run across the sequence of PSG epochs, where multiple predictions for an epoch will be achieved. Then, we calculate the mean probability to predict the correct sleep stage.

 \vspace{-0.3cm}
\subsection{Interpretability}
\label{subsec:interpret}
In this subsection, we propose a method to interpret the predictions of cross-modal transformer, which have the potential to help sleep experts and enhance their confident in the system. The main advantage of our architecture is that the attention mechanism can be simply utilized to learn and interpret: 1) intra-modal relationships, 2) cross-modal relationships and 3) inter-epoch relationships.

We extract the outputs of the intra-modal attention block and calculate the scaled dot-product attention between the representation corresponding to the $CLS_c$ token and other representations corresponding to the non-overlapping windows of the raw signal ($Z_i$), as given in
\begin{equation}
    A_i = \mathrm{softmax}\left(\frac{CLS_c\times Z_i}{\sqrt{d_E}}\right),
    \label{eq:atten}
\end{equation}
to interpret intra-modal relationships. Here, $A_i$ represents the attention score for the $i^{\text{th}}$ time step in a $30$ s PSG epoch, where $i = 1,2,\ldots,60$. Calculating the scaled dot product allows for the identification of time steps in a PSG epoch that are more aligned with the $CLS_c$ token and have a greater impact on the final classification. This gives attention weights for each non-overlapping window of the raw signal, which interpret their impact on the prediction.  

In a similar procedure, the relationships between modalities is interpreted by the scaled dot-product attention of the representations in the output of the cross-modal attention block corresponding to $CLS_c$ of each modality. Finally, we interpret the inter-epoch relationships by calculating the scaled dot-product attention between $CLS_{cross}$ of each epoch.





\begin{table*}[t]
\centering
{
\caption{{\bf Performance comparison between cross-modal transformer and previous works on sleep-EDF-expanded 2018 dataset.}}

\begin{tabular}{l|c|c|c|c|c|c|c|c|c}
\hline  

\multirow{2}{*}{\textbf{ {Method}}}  & \multirow{2}{*}{\textbf{Channels}} & \textbf{Epochs in } & \textbf{Number of} & \multicolumn{6}{c}{\textbf{ {Overall Performance}}} \\ \cline{5-10}

& & \textbf{a Sequence}& \textbf{Parameters} &  {ACC} & $\kappa$ &  {MF1} &  {Sens.} &  {Spec.} &  {MGm}\\
\hline
 
 {SleepEEGNet\cite{mousavi2019}}  &  {Fpz-Cz} &  $10$ & $\sim2.6M$ & 80.0 & 0.730 & 73.6 & $-$ & $-$ & $-$   \\

 {DeepSleepNet\cite{deepsleepnetlite}}  &  {Fpz-Cz} & $25$ & $\sim24.7M$ &  77.1 & 0.69 & 71.2 & $-$ & $-$ & $-$ \\


 {MultitaskCNN\cite{eldele2021}}  &  {Fpz-Cz} & $-$ & $-$ & 79.6 & 0.72 & 72.8 & $-$ & $-$ & 82.5  \\
 
 {AttnSleep\cite{eldele2021}}  &  {Fpz-Cz} & $3$ & $-$ & 82.9 & 0.77 & 78.1 & $-$ & $-$ & 85.6 \\

 {SleepTransformer\cite{Trans1}} &  {Fpz-Cz} & $21$ & $\sim3.7M$ & 81.4 & 0.743 & 74.3 & 74.5 & 95.0 & $84.13$  \\

 {SeqSleepNet\cite{deepsleepnetlite}}  &  {Fpz-Cz} & $20$ & $\sim0.2M$ & 82.6 & 0.76 & 76.4 & 76.3 & 95.4 &$85.32$ \\

 {FCNN+RNN\cite{xsleepnet}}  &  {Fpz-Cz} & $20$ & $\sim5.6M$ & 82.8 & 0.761 & 76.6 & 75.9 & 95.4 &$85.09$\\

 {XSleepNet2\cite{xsleepnet}} &  {Fpz-Cz} & $20$ & $\sim5.8M$ & 84.0 & 0.778 & 77.9 & 77.5 & 95.7 & $86.12$  \\

 {XSleepNet1\cite{xsleepnet}} &  {Fpz-Cz} & $20$ & $\sim5.8M$ & 83.6 & 0.773 & 77.8 & 77.7 & 95.7 & $86.23$ \\

 {NaiveFusion\cite{xsleepnet}} &  {Fpz-Cz} & $20$ & $\sim5.8M$ & 82.3 & 0.755 & 76.2 & 75.7 & 95.3 & $84.94$ \\
    
 {TinySleepNet\cite{deepsleepnetlite}} &  {Fpz-Cz} & $15$ & $\sim1.3M$ & 83.1 & 0.77 & 78.1 & $-$ & $-$ & $-$ \\

SleePyCo without CL\cite{lee2024sleepyco} &  {Fpz-Cz} & $10$& $\sim2.37M$ & 83.5   & 0.772 & 77.7 &  $-$ & $-$  &  $-$\\

SleePyCo*\cite{lee2024sleepyco} &  {Fpz-Cz} & $10$& $\sim2.37M$ & 84.6  & 0.787 & 79.0 &  $-$ & $-$  &  $-$\\

\hline


 {XSleepNet2\cite{xsleepnet}} &  {Fpz-Cz \& EOG} & $20$ & $-$ & {84.0} & {0.778} & {78.7} & 77.6 & 95.7 & $86.18$  \\

 {XSleepNet1\cite{xsleepnet}} &  {Fpz-Cz \& EOG} & $20$ & $-$ & 84.0 & 0.777 & 78.4 & 77.1 & 95.6 & $85.85$ \\

 {SeqSleepNet\cite{xsleepnet}}  &  {Fpz-Cz \& EOG} & $20$ & $-$ & 83.8 & 0.776 & 78.2 & 77.4 & 95.6 &$86.02$ \\

 {FCNN+RNN\cite{xsleepnet}}  &  {Fpz-Cz \& EOG} & $20$ & $-$ & 82.7 & 0.759 & 76.9 & 75.5 & 95.3 &$84.82$\\

 {NaiveFusion\cite{xsleepnet}} &  {Fpz-Cz \& EOG} & $20$ & $-$ & 82.5 & 0.757 & 76.9 & 75.8 & 95.3 & $84.99$ \\

{SleepViTransformer*\cite{peng2023sleepvitransformer}} &  {Fpz-Cz \& EOG} & $4$& $-$ & \textcolor{red}{\bf 85.0}   & \textcolor{red}{\bf 0.792} & 79.1 & $-$ & $-$  & $-$ \\

\hline

\textbf{ {Epoch Cross-}}  &  \multirow{2}{*}{Fpz-Cz \& EOG} & \multirow{2}{*}{$1$} & \multirow{2}{*}{$\sim0.32M$} & 
\multirow{2}{*}{80.8$\pm$1.1}	& \multirow{2}{*}{0.736$\pm$0.014} & \multirow{2}{*}{74.7} & 	\multirow{2}{*}{75.4}	& \multirow{2}{*}{95.0} & \multirow{2}{*}{84.6}   \\
\textbf{ {Modal Transformer}} &    &&& & & & & & \\
\hline
\textbf{ {Sequence Cross-}}  &  \multirow{2}{*}{Fpz-Cz \& EOG} & \multirow{2}{*}{$5$} & \multirow{2}{*}{$\sim1.42M$} &
\multirow{2}{*}{83.7$\pm$1.2}	& \multirow{2}{*}{0.776$\pm$0.016} & \multirow{2}{*}{78.4} & 	\multirow{2}{*}{{79.4}}	& \multirow{2}{*}{{95.8}} & \multirow{2}{*}{{87.2}}   \\
\textbf{ {Modal Transformer - 5}} &   && & & & & & &  \\
\hline
\textbf{ {Sequence Cross-}}  &  \multirow{2}{*}{Fpz-Cz \& EOG} & \multirow{2}{*}{$10$} & \multirow{2}{*}{$\sim2.77M$} &
\multirow{2}{*}{84.0$\pm$1.3}	& \multirow{2}{*}{0.781$\pm$0.017} & \multirow{2}{*}{79.0} & 	\multirow{2}{*}{{80.1}}	& \multirow{2}{*}{{95.8}} & \multirow{2}{*}{{87.6}}   \\
\textbf{ {Modal Transformer - 10}} &   && & & & & & &  \\
\hline
\textbf{ {Sequence Cross-}}  &  \multirow{2}{*}{Fpz-Cz \& EOG} & \multirow{2}{*}{$15$} & \multirow{2}{*}{$\sim4.05M$} &
\multirow{2}{*}{\bf {84.3$\pm$1.3}}	& \multirow{2}{*}{\bf {0.785$\pm$0.017}} & \multirow{2}{*}{\textcolor{red}{\bf 79.4}} & 	\multirow{2}{*}{\textcolor{red}{\bf 80.2}}	& \multirow{2}{*}{\textcolor{red}{\bf 95.9}} & \multirow{2}{*}{\textcolor{red}{\bf 87.7}}   \\
\textbf{ {Modal Transformer - 15}} &   && & & & & & &  \\
\hline
\textbf{ {Sequence Cross-}}  &  \multirow{2}{*}{Fpz-Cz \& EOG} & \multirow{2}{*}{$21$} & \multirow{2}{*}{$\sim5.59M$} &
\multirow{2}{*}{84.1$\pm$1.3}	& \multirow{2}{*}{0.781$\pm$0.017} & \multirow{2}{*}{79.2} & 	\multirow{2}{*}{{80.1}}	& \multirow{2}{*}{{95.9}} & \multirow{2}{*}{{87.6}}   \\
\textbf{ {Modal Transformer - 21}} &   && & & & & & &  \\

\hline


\end{tabular}

\label{table:results}}
\begin{flushright} * refers to the methods that utilize self-supervised pertaining before supervised training for classification. CL indicates contrastive learning.
\end{flushright}
\end{table*}


\begin{table*}[thpb]
\centering
{
\vspace{-0.5cm}
\caption{{\bf Performance comparison between cross-modal transformer and previous works on SHHS dataset.}}

\begin{tabular}{l|c|c|c|c|c|c|c|c|c}
\hline  

\multirow{2}{*}{\textbf{ {Method}}}  & \multirow{2}{*}{\textbf{Channels}} & \textbf{Epochs in } & \textbf{Number of} & \multicolumn{6}{c}{\textbf{ {Overall Performance}}} \\ \cline{5-10}

& & \textbf{a Sequence}& \textbf{Parameters} &  {ACC} & $\kappa$ &  {MF1} &  {Sens.} &  {Spec.} &  {MGm}\\
\hline
 



 {MultitaskCNN\cite{eldele2021}}  &  {C4-A1} & $-$ & $-$& 81.4   &0.74  & 71.2 & $-$ & $-$ &  83.6 \\
 
 {AttnSleep\cite{eldele2021}}  &  {C4-A1} & $3$ & $-$&  84.2 &  0.78& 75.3 & $-$ & $-$ &  84.0\\

 {SleepTransformer\cite{Trans1}} &  {C4-A1} & $21$ & $\sim3.7M$ &  87.7 & 0.828 & 80.1 & 78.7 & 96.5 & 87.15  \\

 {SeqSleepNet\cite{xsleepnet}}  &  {C4-A1} & $20$& $\sim0.2M$ &  86.5 & 0.811 & 78.5 & 76.9 & 96.1 & 85.97\\

 {FCNN+RNN\cite{xsleepnet}}  &  {C4-A1} & $20$& $\sim5.6M$ &  86.7 & 0.813 & 79.5  & 78.1  & 96.2  & 86.68  \\

 {XSleepNet2\cite{xsleepnet}} &  {C4-A1} & $20$& $\sim5.8M$ & 87.6  & 0.826  & 80.7  & 79.7 & 96.5  & 87.7  \\

 {XSleepNet1\cite{xsleepnet}} &  {C4-A1} & $20$& $\sim5.8M$ &  87.5  & 0.826 & 81.0  & 80.4  & 96.5 &  88.1 \\

 {NaiveFusion\cite{xsleepnet}} &  {C4-A1} & $20$& $\sim5.8M$ & 87.5   & 0.825  & 80.7  & 79.8  & 96.5  & 87.8  \\

  {SleePyCo*\cite{lee2024sleepyco}} &  {C4-A1} & $10$& $\sim2.37M$ & 87.9   & 0.830 & 80.7 &  $-$ & $-$  &  $-$\\

  {CoRe-Sleep\cite{kontras2024core}} &  {C4-A1} & $-$& $-$ & {88.2}   & {0.834} & {80.8} &  $-$ & $-$  &  $-$\\


\hline


 {XSleepNet2\cite{xsleepnet}} &  {C4-A1 \& EOG} & $20$& $-$ & {88.8}  & {0.843}  & {81.8}  & 80.8 & {96.8}  & 88.4  \\

 {XSleepNet1\cite{xsleepnet}} &  {C4-A1 \& EOG} & $20$& $-$ & 88.8  & 0.843 & 82.0  & 81.3  & 96.8 & 88.7 \\

 {SeqSleepNet\cite{xsleepnet}}  &  {C4-A1 \& EOG} & $20$& $-$ & 88.4 & 0.837 & 80.7 & 79.6 & 96.7 & 87.7\\

 {FCNN+RNN\cite{xsleepnet}}  &  {C4-A1 \& EOG} & $20$& $-$ & 88.0  & 0.831 & 80.5 & 79.2  & 96.6&87.47\\

 {NaiveFusion\cite{xsleepnet}} &  {C4-A1 \& EOG} & $20$& $-$ & 88.4   & 0.839 & 81.7  & 81.6  & 96.8  & 88.9  \\

 {SleepViTransformer*\cite{peng2023sleepvitransformer}} &  {C4-A1 \& EOG} & $4$& $-$ & 88.1   & 0.83 & 79.8  & $-$ & $-$  & $-$ \\

{CoRe-Sleep\cite{kontras2024core}} &  {C4-A1 \& EOG} & $-$& $-$ & \textcolor{red}{\bf89.5}   & \textcolor{red}{\bf0.853} & \textcolor{red}{\bf82.3} &  $-$ & $-$  &  $-$\\
 

\hline


\textbf{ {Epoch Cross-}}  &  \multirow{2}{*}{C4-A1 \& EOG} & \multirow{2}{*}{$1$} & \multirow{2}{*}{$\sim0.32M$} & 
\multirow{2}{*}{79.3}	& \multirow{2}{*}{0.7116} & \multirow{2}{*}{67.6} & 	\multirow{2}{*}{{68.1}}	& \multirow{2}{*}{{94.4}} & \multirow{2}{*}{{80.2}}   \\
\textbf{ {Modal Transformer}} &   &&  & & & & &  \\
\hline

\textbf{ {Sequence Cross-}}  &  \multirow{2}{*}{C4-A1 \& EOG} & \multirow{2}{*}{$5$} & \multirow{2}{*}{$\sim1.42M$} & 
\multirow{2}{*}{87.5}	& \multirow{2}{*}{0.827} & \multirow{2}{*}{80.9} & 	\multirow{2}{*}{{81.9}}	& \multirow{2}{*}{{96.6}} & \multirow{2}{*}{{88.9}}   \\
\textbf{ {Modal Transformer - 5}} &   &&  & & & & &  \\
\hline
\textbf{ {Sequence Cross-}}  &  \multirow{2}{*}{C4-A1 \& EOG} & \multirow{2}{*}{$10$} & \multirow{2}{*}{$\sim2.77M$} & 
\multirow{2}{*}{\bf{87.7}}	& \multirow{2}{*}{\bf {0.829}} & \multirow{2}{*}{{81.4}} & 	\multirow{2}{*}{82.6}	& \multirow{2}{*}{ {96.7}} & \multirow{2}{*}{{89.4}}   \\
\textbf{ {Modal Transformer - 10}} &   & & & & & & &  \\
\hline

\textbf{ {Sequence Cross-}}  &  \multirow{2}{*}{C4-A1 \& EOG} & \multirow{2}{*}{$15$} & \multirow{2}{*}{$\sim4.05M$} & 
\multirow{2}{*}{{87.6}}	& \multirow{2}{*}{{0.828}} & \multirow{2}{*}{{\bf81.5}} & 	\multirow{2}{*}{\textcolor{red}{\bf 83.1}}	& \multirow{2}{*}{{\bf96.7}} & \multirow{2}{*}{{89.6}}   \\
\textbf{ {Modal Transformer - 15}} &   &&  & & & & &  \\
\hline

\textbf{ {Sequence Cross-}}  &  \multirow{2}{*}{C4-A1 \& EOG} & \multirow{2}{*}{$21$} & \multirow{2}{*}{$\sim5.59M$} & 
\multirow{2}{*}{87.6}	& \multirow{2}{*}{0.827} & \multirow{2}{*}{81.4} & 	\multirow{2}{*}{{83.2}}	& \multirow{2}{*}{{96.7}} & \multirow{2}{*}{\textcolor{red}{\bf89.7}}   \\
\textbf{ {Modal Transformer - 21}} &   &&  & & & & &  \\

\hline


\end{tabular}
\label{table:shhs_results}}
\begin{flushright} * refers to the methods that utilize self-supervised pertaining before supervised training for classification.
\end{flushright}
\end{table*}


\vspace{-0.2cm}
\section{Experiments}
\label{sec:Experiments}
\subsection{Dataset}
\label {sec:Dataset}
\subsubsection{SleepEDF expanded dataset:} We used the publicly available sleep-EDF-expanded dataset (Sleep-EDF-78)~\cite{kemp2000} from Physionet~\cite{goldberger2000} to evaluate the proposed architecture for sleep stage classification. We used the sleep cassette (SC) dataset of sleep-EDF 2018 for our experiments, which consists of $153$ whole night PSG recordings from $78$ healthy individuals. 
Each recording comprises of two bipolar EEG channels (Fpz-Cz and Pz-Oz), a horizontal EOG signal and a submental chin EMG signal. For our experiments, we only utilized Fpz-Cz EEG channel and EOG signal. The PSG recordings were accompanied with hypnograms annotated by sleep technicians based on Rechtschaffen and Kales (R\&K) guidelines~\cite{rechtschaffen1968manual}. In the hypnogram, each $30$ s epoch of the recorded data is assigned to one of the following labels : Stage $0$, Stage $1$, Stage $2$, Stage $3$, Stage $4$, REM, movement time and `?' (unscored). For this study, we converted these annotations to the AASM standards~\cite{Iber2007}, by combining Stage $3$ and Stage $4$ to a single stage N3 while Stage $0$, Stage $1$ and Stage $2$ are relabelled as Wake, N1 and N2, respectively. Also, epochs consisting of annotations `movement time' and `?' were discarded. Altogether $415,465$ epochs of $30$s duration of PSG were extracted. We employed a five-fold cross validation method, where the dataset was divided into subject-independent groups to ensure a fair evaluation of the model performance (Subjects in each group are independent.).

\subsubsection{SHHS dataset:} We also utilized the publicly available Sleep Heart Health Study (SHHS) dataset \cite{zhang2018national,quan1997sleep} in our research. SHHS is a multicenter, longitudinal study examining the effects of sleep-disordered breathing on cardiovascular diseases. This dataset includes two sets of recordings: SHHS1 (visit 1) and SHHS2 (visit 2), where we used the SHHS1 (visit 1) for our study. Each recording contains two EEG channels (C4-A1 and C3-A2), and we specifically used the C4-A1 EEG channel along with EOG in our experiments similar to past works. SHHS was initially annotated according to R\&K guidelines. We converted these annotations to AASM using the same approach applied to sleepEDF. The epochs containing MOVEMENT and UNKNOWN were discarded, and recordings without all sleep stages were also discarded. Following \cite{Trans1}, we randomly split the subjects in the dataset into 70\% for training and 30\% for testing, with 100 subjects from the training set separated as a validation set.

The data processing for sleepEDF and the training setups are detailed in the supplementary materials (Section A and B).

\begin{table}[!t]
\centering
{
\caption{{\bf Class-wise performance comparison on sleep-EDF-expanded 2018 dataset.}}

\resizebox{\columnwidth}{!}{%
\begin{tabular}{l|c|c|c|c|c|c}
\hline  

\multirow{2}{*}{\textbf{ {Method}}}  & \multirow{2}{*}{\textbf{\footnotesize{Chan.}}} &  \multicolumn{5}{c}{\textbf{ {Per-class Performance}}} \\ \cline{3-7}

& &  {W}&  {N1} &  {N2} &  {N3} &  {REM} \\
\hline
 
 {SleepEEGNet\cite{mousavi2019}}  &  {Fpz-Cz} &  91.7  &  44.1  &  82.5  &  73.5  &  76.1  \\

 {DeepSleepNet\cite{deepsleepnetlite}}  &  {Fpz-Cz} &  90.4  &  46.0  &  79.1  &  68.6  &  71.8  \\


 {MultitaskCNN\cite{eldele2021}}  &  {Fpz-Cz} &  90.9  &  39.7  &  83.2  &  76.6  &  73.5   \\
 
 {AttnSleep\cite{eldele2021}}  &  {Fpz-Cz} &  92.6  &  47.4  &  85.5  & \textcolor{red}{83.7} &  81.5   \\

 {SleepTransformer\cite{Trans1}} &  {Fpz-Cz} &  91.7  &  40.4  &  84.3  &  77.9  &  77.2     \\

 {SeqSleepNet\cite{deepsleepnetlite}}  &  {Fpz-Cz} &  91.8  &  42.6  & \textcolor{red}{86.5} &  76.4  &  84.1  \\

 {FCNN+RNN\cite{xsleepnet}}  &  {Fpz-Cz} &  92.5  &  47.3  &  85.0  &  79.2  &  78.9 \\

 {XSleepNet2\cite{xsleepnet}} &  {Fpz-Cz} &  93.3  &  49.9  &  86.0  &  78.7  &  81.8  \\

 {XSleepNet1\cite{xsleepnet}} &  {Fpz-Cz} &  92.6  &  50.2  &  85.9  &  79.2  &  81.3  \\

 {NaiveFusion\cite{xsleepnet}} &  {Fpz-Cz} &  93.2  &  49.6  &  86.2  &  79.4  &  82.5 \\

 {TinySleepNet\cite{deepsleepnetlite}} &  {Fpz-Cz} &  92.8  &  51.0  &  85.3  &  81.1  &  80.3  \\
 
SleePyCo  &  \multirow{2}{*}{Fpz-Cz} &\multirow{2}{*}{93.2}& \multirow{2}{*}{47.9} & \multirow{2}{*}{85.1} &\multirow{2}{*}{79.9} & \multirow{2}{*}{82.3} \\ without CL\cite{lee2024sleepyco} &&&&&&\\

SleePyCo*\cite{lee2024sleepyco} &  {Fpz-Cz} &  93.5  &  50.4  &  86.5  &  80.5  &  84.2  \\
 
\hline
\multirow{2}{*}{FCNN + RNN \cite{xsleepnet}}  &  {Fpz-Cz}  & 
\multirow{2}{*}{91.2}	& \multirow{2}{*}{45.8} & \multirow{2}{*}{84.7} & 	\multirow{2}{*}{78.5}	& \multirow{2}{*}{84.2} \\ &  {EOG} & & & & & \\

\multirow{2}{*}{XSleepNet2 \cite{xsleepnet}}  &  {Fpz-Cz}  & 
\multirow{2}{*}{92.6}	& \multirow{2}{*}{50.3} & \multirow{2}{*}{85.5} & 	\multirow{2}{*}{79.2}	& \multirow{2}{*}{85.7} \\ &  {EOG} & & & & & \\

\multirow{2}{*}{XSleepNet1 \cite{xsleepnet}}  &  {Fpz-Cz}  & 
\multirow{2}{*}{92.2}	& \multirow{2}{*}{49.1} & \multirow{2}{*}{85.6} & 	\multirow{2}{*}{78.8}	& \multirow{2}{*}{86.3} \\ &  {EOG} & & & & & \\

\multirow{2}{*}{NaiveFusion \cite{xsleepnet}}  &  {Fpz-Cz}  & 
\multirow{2}{*}{91.0}	& \multirow{2}{*}{47.7} & \multirow{2}{*}{84.7} & 	\multirow{2}{*}{77.7}	& \multirow{2}{*}{83.6} \\ &  {EOG} & & & & & \\

SleepVi-\cite{peng2023sleepvitransformer} &  {Fpz-Cz}  & 
\multirow{2}{*}{93.6}	& \multirow{2}{*}{49.4} & \multirow{2}{*}{86.4} & 	\multirow{2}{*}{79.3}	& \multirow{2}{*}{\textcolor{red}{86.9}} \\Transformer* &  {EOG} & & & & & \\
 \hline
 
 {Epoch Cross-}  &  {Fpz-Cz}  & 
\multirow{2}{*}{92.3}	& \multirow{2}{*}{45.3} & \multirow{2}{*}{82.9} & 	\multirow{2}{*}{75.5}	& \multirow{2}{*}{77.4}   \\
 {Modal Transformer} &   {EOG} & & & & & \\
\hline
{Sequence Cross-}  &  {Fpz-Cz} & 
\multirow{2}{*}{{ \textcolor{red}{93.7}}}	& \multirow{2}{*}{{52.4}} & \multirow{2}{*}{85.0} & 	\multirow{2}{*}{76.0}	& \multirow{2}{*}{85.3}   \\
{Modal Transformer-5} &    {EOG}& & & & &  \\
\hline

{Sequence Cross-}  &  {Fpz-Cz} & 
\multirow{2}{*}{{93.4}}	& \multirow{2}{*}{{53.6}} & \multirow{2}{*}{85.1} & 	\multirow{2}{*}{76.4}	& \multirow{2}{*}{86.6}   \\
{Modal Transformer-10} &    {EOG}& & & & &  \\
\hline

{Sequence Cross-}  &  {Fpz-Cz} & 
\multirow{2}{*}{{93.5}}	& \multirow{2}{*}{ \textcolor{red}{54.3}} & \multirow{2}{*}{85.5} & 	\multirow{2}{*}{77.0}	& \multirow{2}{*}{ 86.7}   \\
{Modal Transformer-15} &    {EOG}& & & & &  \\
\hline

{Sequence Cross-}  &  {Fpz-Cz} & 
\multirow{2}{*}{{93.5}}	& \multirow{2}{*}{ \textcolor{red}{{54.3}}} & \multirow{2}{*}{85.0} & 	\multirow{2}{*}{76.9}	& \multirow{2}{*}{86.4}   \\
{Modal Transformer-21} &    {EOG}& & & & &  \\

\hline
\end{tabular}}
\begin{flushright} * \scriptsize indicates methods using self-supervised pre-training before supervised training for classification. CL denotes contrastive learning.
\end{flushright}
\vspace{-0.9cm}
\label{table: class_per}}
\end{table}


\vspace{-0.1cm}
\section{Results and Discussion}
\label{sec:Results}

\subsection{Sleep Stage Classification}
We evaluated the performance of our proposed epoch and sequence cross-modal transformers on the SleepEDF-expanded 2018 and SHHS datasets, comparing it against various state-of-the-art methods. The performance metrics used are accuracy (ACC), Cohen's kappa ($\kappa$), macro-F1 score (MF1), sensitivity (Sens), specificity (Spec), and macro-averaged G-mean (MGm). Further details on the performance metrics are provided in the supplementary materials (Section C). The results of the methods for both classification schemes (one-to-one and many-to-many classification) are presented in Tables \ref{table:results} and \ref{table:shhs_results}. Additionally, we evaluate the class-wise performance, with results provided in Tables \ref{table: class_per} and \ref{table: shhs_class_per}.

In the SleepEDF dataset, our proposed sequence cross-modal transformer with $15$ epochs in a sequence achieves comparable performance with the state-of-the-art methods for accuracy and Cohen's Kappa coefficient ($\kappa$). However, our method outperforms past works in macro-averaged F1 ($79.4\%$), sensitivity ($80.2\%$), specificity ($95.9\%$), and macro-averaged G-mean ($87.7\%$). Similarly, the same sequence cross-modal transformer outperforms past works in sensitivity ($83.1\%$) and macro-averaged G-mean ($89.6\%$) in the SHHS dataset. For other metrics, our performance is slightly lower than past works. A key factor to note is that our sequence cross-modal transformer achieves high classification performance with fewer parameters compared to previous methods. While SeqSleepNet \cite{phan2019seqsleepnet} has fewer parameters, it relies on RNNs, which require longer training and inference times.

Compared to XSleepNets \cite{xsleepnet}, our sequence cross-modal transformer achieves better performance with fewer PSG epochs in the sequence ($5$, $10$, and $15$) in the SleepEDF dataset, though it performs slightly lower in the SHHS dataset. Despite this, our method offers better interpretability and smaller model print compared to XSleepNets. {CoRe-Sleep\cite{kontras2024core} achieves better performance compared to ours, however the method does not focus on interpretability.} Furthermore, we compared the performance with existing transformer-based methods SleepTransformer \cite{Trans1} and SleepViTransformer \cite{peng2023sleepvitransformer}. Without any pretraining on a larger database, our sequence cross-modal transformer outperforms SleepTransformer on the SleepEDF dataset, demonstrating its modeling capability on smaller datasets. SleepViTransformer benefits from cross-modality pretraining using image and audio data, significantly enhancing its performance by $4\%$ in accuracy based on their study on the SleepEDF-20 dataset \cite{peng2023sleepvitransformer}. Our method achieves comparable performance to SleepViTransformer and a higher macro-F1 score on the SleepEDF-78 and SHHS datasets without using external data or pretraining. Our method also outperforms SleePyCo \cite{lee2024sleepyco} without their contrastive learning strategy and achieves a better macro-F1 score compared to SleePyCo with contrastive learning. We strongly believe that deep learning techniques such as transfer learning, meta learning~\cite{meta_2021}, knowledge distillation~\cite{KD_2022}, large-scale training and self-supervised pretraining could further improve the performance of our method. In our future studies, we plan to explore self-supervised pretraining strategies to further improve the performance of interpretable deep learning models for sleep staging.

\begin{table}[!t]
\centering
{
\caption{{\bf Class-wise performance comparison on SHHS dataset.}}

\resizebox{\columnwidth}{!}{%
\begin{tabular}{l|c|c|c|c|c|c}
\hline  

\multirow{2}{*}{\textbf{ {Method}}}  & \multirow{2}{*}{\textbf{\footnotesize{Chan.}}} &  \multicolumn{5}{c}{\textbf{ {Per-class Performance}}} \\ \cline{3-7}

& &  {W}&  {N1} &  {N2} &  {N3} &  {REM} \\
\hline

 {MultitaskCNN\cite{eldele2021}}  &  {C4-A1} &  82.2  &  25.7  &  83.9  &  83.3  &  81.1   \\
 
 {AttnSleep\cite{eldele2021}}  &  {C4-A1} &  86.7  &  33.2  &  87.1  &  \textcolor{red}{87.5}  &  82.1   \\

 {SleepTransformer\cite{Trans1}} &  {C4-A1} &  92.2  &  46.1  &  88.3  &  85.2  &  88.6     \\

 {SeqSleepNet\cite{xsleepnet}}  &  {C4-A1} &  91.4  &  43.3  &  87.4  &  82.9  &  87.3  \\

 {FCNN+RNN\cite{xsleepnet}}  &  {C4-A1} &  91.1  &  48.7  &  88.0  &  82.6  &  87.1 \\

 {XSleepNet2\cite{xsleepnet}} &  {C4-A1} &  92.0  &  49.9  &  88.3  &  85.0  &  88.2  \\

 {XSleepNet1\cite{xsleepnet}} &  {C4-A1} &  91.6  &  51.4  &  88.5  &  85.0  &  88.4  \\

 {NaiveFusion\cite{xsleepnet}} &  {C4-A1} &  91.9  &  50.9  &  88.4  &  84.1  &  88.3 \\

 {SleePyCo*\cite{lee2024sleepyco}} &  {C4-A1} &  92.6  &  49.2  &  88.5  &  84.5  &  88.6  \\

\hline

\multirow{2}{*}{XSleepNet2 \cite{xsleepnet}}  &  {C4-A1}  & 
\multirow{2}{*}{93.3}	& \multirow{2}{*}{50.2} & \multirow{2}{*}{\textcolor{red}{89.5}} & 	\multirow{2}{*}{85.1}	& \multirow{2}{*}{91.0} \\ &  {EOG} & & & & & \\

\multirow{2}{*}{XSleepNet1 \cite{xsleepnet}}  &  {C4-A1}  & 
\multirow{2}{*}{\textcolor{red}{93.4}}	& \multirow{2}{*}{51.0} & \multirow{2}{*}{89.4} & 	\multirow{2}{*}{84.9}	& \multirow{2}{*}{\textcolor{red}{91.3}} \\ &  {EOG} & & & & & \\

\multirow{2}{*}{SeqSleepNet \cite{xsleepnet}}  &  {C4-A1}  & 
\multirow{2}{*}{93.1}	& \multirow{2}{*}{46.2} & \multirow{2}{*}{88.9} & 	\multirow{2}{*}{84.7}	& \multirow{2}{*}{90.8} \\ &  {EOG} & & & & & \\

\multirow{2}{*}{FCNN + RNN \cite{xsleepnet}}  &  {C4-A1}  & 
\multirow{2}{*}{92.3}	& \multirow{2}{*}{47.0} & \multirow{2}{*}{88.7} & 	\multirow{2}{*}{84.3}	& \multirow{2}{*}{90.4} \\ &  {EOG} & & & & & \\

\multirow{2}{*}{NaiveFusion \cite{xsleepnet}}  &  {C4-A1}  & 
\multirow{2}{*}{93.0}	& \multirow{2}{*}{50.9} & \multirow{2}{*}{89.2} & 	\multirow{2}{*}{85.2}	& \multirow{2}{*}{90.2} \\ &  {EOG} & & & & & \\

SleepVi-  &  {C4-A1}  & 
\multirow{2}{*}{93.4}	& \multirow{2}{*}{44.4} & \multirow{2}{*}{88.5} & 	\multirow{2}{*}{84.5}	& \multirow{2}{*}{88.3} \\Transformer* \cite{peng2023sleepvitransformer} &  {EOG} & & & & & \\
 \hline
{Epoch Cross-}  &  {C4-A1}  & 
\multirow{2}{*}{89.7}	& \multirow{2}{*}{20.9} & \multirow{2}{*}{80.8} & 	\multirow{2}{*}{75.7}	& \multirow{2}{*}{71.0}   \\
 {Modal Transformer} &   {EOG} & & & & & \\
\hline
{Sequence Cross-}  &  {C4-A1} & 
\multirow{2}{*}{{ 92.7}}	& \multirow{2}{*}{{51.1}} & \multirow{2}{*}{88.2} & 	\multirow{2}{*}{83.7}	& \multirow{2}{*}{89.1}   \\
{Modal Transformer-5} &    {EOG}& & & & &  \\
\hline

{Sequence Cross-}  &  {C4-A1} & 
\multirow{2}{*}{{92.8}}	& \multirow{2}{*}{{52.5}} & \multirow{2}{*}{88.3} & 	\multirow{2}{*}{83.7}	& \multirow{2}{*}{89.8}   \\
{Modal Transformer-10} &    {EOG}& & & & &  \\
\hline

{Sequence Cross-}  &  {C4-A1} & 
\multirow{2}{*}{{92.6}}	& \multirow{2}{*}{ {52.8}} & \multirow{2}{*}{88.3} & 	\multirow{2}{*}{83.5}	& \multirow{2}{*}{90.1}   \\
{Modal Transformer-15} &    {EOG}& & & & &  \\
\hline

{Sequence Cross-}  &  {C4-A1} & 
\multirow{2}{*}{{92.8}}	& \multirow{2}{*}{ \textcolor{red}{{52.9}}} & \multirow{2}{*}{88.1} & 	\multirow{2}{*}{83.4}	& \multirow{2}{*}{90.0}   \\
{Modal Transformer-21} &    {EOG}& & & & &  \\





 
\hline
\end{tabular}}
\begin{flushright} \scriptsize* indicates methods using self-supervised pre-training before supervised training.
\end{flushright}
\vspace{-0.2cm}
\label{table: shhs_class_per}}
\end{table}

When considering class-wise performance based on F1-score in Table \ref{table: class_per} and \ref{table: shhs_class_per}, our sequence cross-modal transformer achieves the state-of-the-art performance in the prediction of Wake and N1 in SleepEDF dataset and N1 in SHHS dataset. This enhancement in the performance can be attributed to the capability of our method to learn cross-modal relationships, where EOG along with EEG makes an impact on their predictions. The performance of our sequence cross-modal transformer in predicting N2, N3 and REM stages are on-par with the previous work.     

\begin{table}[!t]
\vspace{-0.4cm}
\centering
\caption{{\bf Model size and training time comparison}}

\resizebox{\columnwidth}{!}{%
\begin{tabular}{l|c|c}
\hline  

\multirow{2}{*}{\textbf{Method}}&\multirow{2}{*}{\textbf{No of Parameters}} & \textbf{Training time }\\
&& \textbf{per 1000 steps (s)}\\
\hline
 
SleepTransformer\cite{Trans1} &$\sim3.7M$ & 308\\

XSleepNet2\cite{xsleepnet} &$\sim5.8M$ & 828\\

SeqSleepNet\cite{deepsleepnetlite} & $\sim0.2M$& 379\\

\hline

\textbf{Epoch Cross-} &\multirow{2}{*}{$\sim0.32M$}  &  \multirow{2}{*}{53}\\
\textbf{Modal Transformer} & &\\

\textbf{Sequence Cross-}  &\multirow{2}{*}{$\sim1.42M$} &  \multirow{2}{*}{174} \\
\textbf{Modal Transformer - 5} & &\\

\textbf{Sequence Cross-}  &\multirow{2}{*}{$\sim2.77M$} &  \multirow{2}{*}{176} \\
\textbf{Modal Transformer - 10} & &\\

\textbf{Sequence Cross-}  &\multirow{2}{*}{$\sim4.05M$} &  \multirow{2}{*}{243} \\
\textbf{Modal Transformer - 15} & &\\

\textbf{Sequence Cross-}  &\multirow{2}{*}{$\sim5.59M$} &  \multirow{2}{*}{347} \\
\textbf{Modal Transformer - 21} & &\\

\hline
\end{tabular}}
\label{table: computational_results}
\end{table}

\subsection{Comparison of Computational Complexity}

Table \ref{table: computational_results} shows the comparison of model size in number of parameters and training time taken for 1000 steps between the variants of the proposed cross-modal transformers and the previously reported work. Here, our method trains faster compared to current the state-of-the-art XSleepNets because their architecture is based on RNNs which requires to process the data sequentially. Unlike RNNs, transformers are capable of processing data in parallel. The parallelism in the transformers enabled our proposed method to train faster compared to current the state-of-the-art XSleepNets and other previous work as given in Table \ref{table: computational_results}. Our epoch transformer achieves performance (ACC : $80.8$) closer to the SleepTransformer (ACC : $81.4$)  model with significantly smaller model footprint ($11.5$ times smaller) and lower training time. In comparison to XSleepNets, our sequence cross-modal transformer with $10$ epochs per sequence achieves better performance with twofold smaller in size and around $5$ times faster in training.




\vspace{-1em}
\subsection{Ablation Study}
Here, we systematically study the importance and advantages of using EOG, cross-modal attention and $CLS$ vectors as aggregated representation in our method. Additionally, we study the impact of different representation learning methods on the performance and the results are provided in the supplementary materials (Section D).

\subsubsection{Importance of EOG and Cross-Modal Attention}
In manual annotation of sleep stages, EOG signals plays a major role in identifying Wake, N1 and REM sleep stages. This shows the importance of developing an algorithm to utilize the features of EOG signals along with EEG for sleep staging. We show the importance of efficiently utilizing EOG and learning the cross-relationship between EEG and EOG modalities by conducting a separate study. In this study, the epoch cross-modal transformer with the model variants shown in Fig.~\ref{fig:EpochCMTAblation} and their corresponding sequence model variants are used. The results of the study are given in Table \ref{table: ablation_study_2}. The proposed model architecture significantly contributes to the performance, as it can be inferred from the performance increase of $1.8\%$ achieved in accuracy by a only EEG variant of sequence cross-modal transformer compared to the SleepTransformer\cite{Trans1}. Also, our single channel EEG model outperforms majority of the existing EEG-only models and it is on-par with the XSleepnets\cite{xsleepnet}.

As anticipated, the model using both EEG and EOG signals had better performance than only EEG, which clearly states that our method is learning from both EEG and EOG signals efficiently. Significant improvement in the prediction of N1 and REM sleep stages can be observed when EOG is added into our method. In order to identify the effective representation learning frameworks two strategies were analysed: (a) learning joint features from concatenated EEG and EOG channels using a single multi-scale 1D-CNN block (EEG + EOG (Jt)) and (b) learn separate features for EEG and EOG using separate multi-scale 1D-CNN blocks (EEG + EOG). As per the results in Table \ref{table: ablation_study_2}, having two separated multi-scale 1D-CNN blocks achieves better performance.  Performance was further improved when the cross-modal attention was incorporated into our method. Inclusion of cross-modal attention significantly improved the prediction of Wake, N2, N3 and REM stages. 
\begin{figure*}[!t]
 \centering
    \includegraphics[width = 0.78 \linewidth]{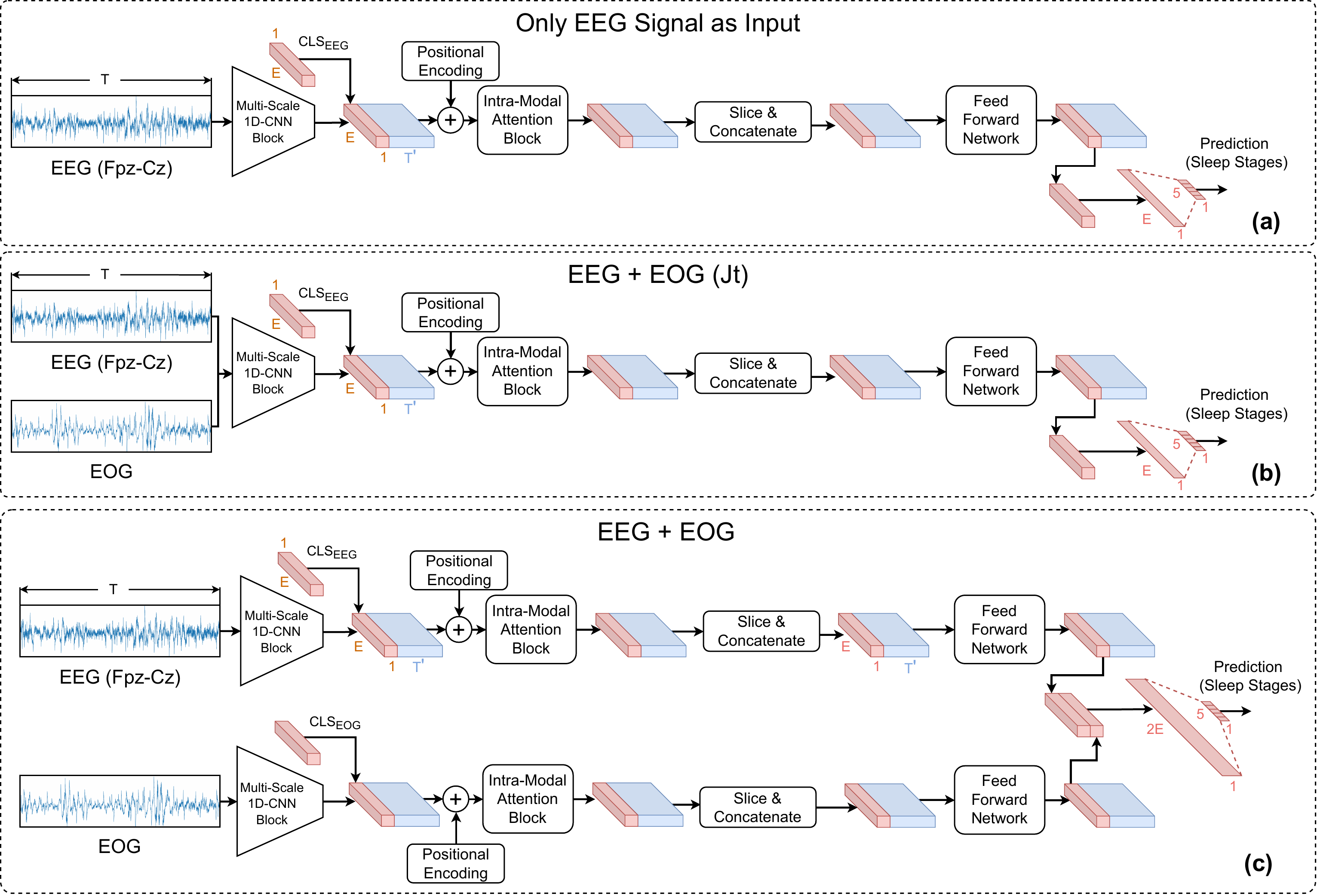}
\caption{The variants of the epoch level model used to conduct the ablation study. (a) shows the architecture with single channel EEG as the input, (b) shows the extended version of (a) with both EEG and EOG as inputs, and (c) shows a version of the epoch cross-modal transformer without cross-modal attention.} 

\label{fig:EpochCMTAblation}
\vspace{-0.5cm}
\end{figure*}

\subsubsection{Importance of using $CLS$}
In our method, utilizing $CLS$ vectors as aggregated representation enabled the reduction of the model size. The size of our initial version of epoch cross-modal transformer without having $CLS$ was $\sim2.1M$, which is $6.5$ times larger than our proposed epoch cross-modal transformer. Addition of $CLS$ vector enabled scaling of the proposed methods in terms of number of modalities and number of epochs in a sequence. Also, the cross-relationships between any number of modalities can be learned easily because we leverage a simple $CLS$ latent vector for each modality which aggregates all the sequence information. Most importantly, the $CLS$ vectors enabled interpretations of the predictions by employing self-attention, which is simple and effective. 

\begin{table}[!t]
\centering
\caption{{\bf The results of the study on the importance of EOG and cross-modal relationships for sleep staging on sleep-EDF-expanded 2018 dataset.}}

\resizebox{0.97\columnwidth}{!}{%
\begin{tabular}{c|c|c|c|c|c|c|c}
\hline  

\multirow{2}{*}{\textbf{\footnotesize Method}} & \multirow{2}{*}{\textbf{\footnotesize ACC}} & \multirow{2}{*}{\textbf{$\kappa$}} & \multicolumn{5}{c}{\textbf{ {\footnotesize Per-class Performance}}} \\ \cline{4-8}
& &&  {\scriptsize W}&  {\scriptsize N1} &  {\scriptsize N2} &  {\scriptsize N3} &  {\scriptsize REM} \\
\hline

\multicolumn{8}{c}{\textbf{\footnotesize Epoch Cross-Modal Transformer}}\\
\hline

\scriptsize{Only EEG} & 78.3 & 0.704 & 91.4 & 37.7 & 81.6 & 75.3 & 69.3 \\

\scriptsize{ EEG + EOG (Jt)} & \footnotesize{80.0} & \footnotesize{0.727} & \footnotesize{92.1} & \footnotesize{\textcolor{red}{45.5}} & \footnotesize{81.9} & \footnotesize{74.6} & \footnotesize{76.7} \\

\scriptsize{EEG + EOG} & 80.4 & 0.731 & \textcolor{red}{92.4} & 44.1 & 82.4 & 74.5 & 76.8 \\ 

\scriptsize{EEG + EOG} +  & \multirow{2}{*}{\textcolor{red}{80.8}} & \multirow{2}{*}{\textcolor{red}{0.736}} & \multirow{2}{*}{92.3} & \multirow{2}{*}{{45.3}} & \multirow{2}{*}{\textcolor{red}{82.9}} & \multirow{2}{*}{\textcolor{red}{75.5}} & \multirow{2}{*}{\textcolor{red}{77.4}} \\
\scriptsize{CMA (Proposed)} & &&&&&&\\

\hline
\multicolumn{8}{c}{\textbf{\footnotesize Sequence Cross-Modal Transformer - 15}}\\
\hline

\scriptsize{Only EEG} & \footnotesize{83.2} & \footnotesize{0.770} & \footnotesize{92.9} & \footnotesize{51.6} & \footnotesize{84.8} & \footnotesize{75.4} & \footnotesize{83.5} \\

\scriptsize{ EEG + EOG (Jt)} & \footnotesize{84.1} & \footnotesize{0.783} & \footnotesize{93.2} & \footnotesize{\textcolor{red}{55.0}} & \footnotesize{85.3} & \footnotesize{77.0} & \footnotesize{86.4} \\

\scriptsize{EEG + EOG} &\footnotesize{84.2} & \footnotesize{0.783} & \footnotesize{93.6} & \footnotesize{53.9} & \footnotesize{85.3} & \footnotesize{76.9} & \footnotesize{86.6} \\

\scriptsize{EEG + EOG +}  & \multirow{2}{*}{\textcolor{red}{84.3}} & \multirow{2}{*}{\textcolor{red}{0.785}} & \multirow{2}{*}{\textcolor{red}{93.5}} & \multirow{2}{*}{{54.3}} & \multirow{2}{*}{\textcolor{red}{85.5}} & \multirow{2}{*}{\textcolor{red}{77.0}} & \multirow{2}{*}{\textcolor{red}{86.7}} \\
\scriptsize{CMA (Proposed)} & &&&&&&\\

\hline
\end{tabular}}
\begin{flushleft} *CMA refers to cross-modal attention and Jt refers to joint features from multi-scale 1D-CNN for combined EEG and EOG as inputs. The model architecture variants for only EEG, EEG + EOG (Jt) and EEG + EOG are illustrated in Fig.~\ref{fig:EpochCMTAblation}(a), Fig.~\ref{fig:EpochCMTAblation}(b) and Fig.~\ref{fig:EpochCMTAblation}(c), respectively.
\end{flushleft}
\label{table: ablation_study_2}
\vspace{-0.3cm}
\end{table}

\begin{figure}[!t]
 \centering
     \includegraphics[width = 1\linewidth]{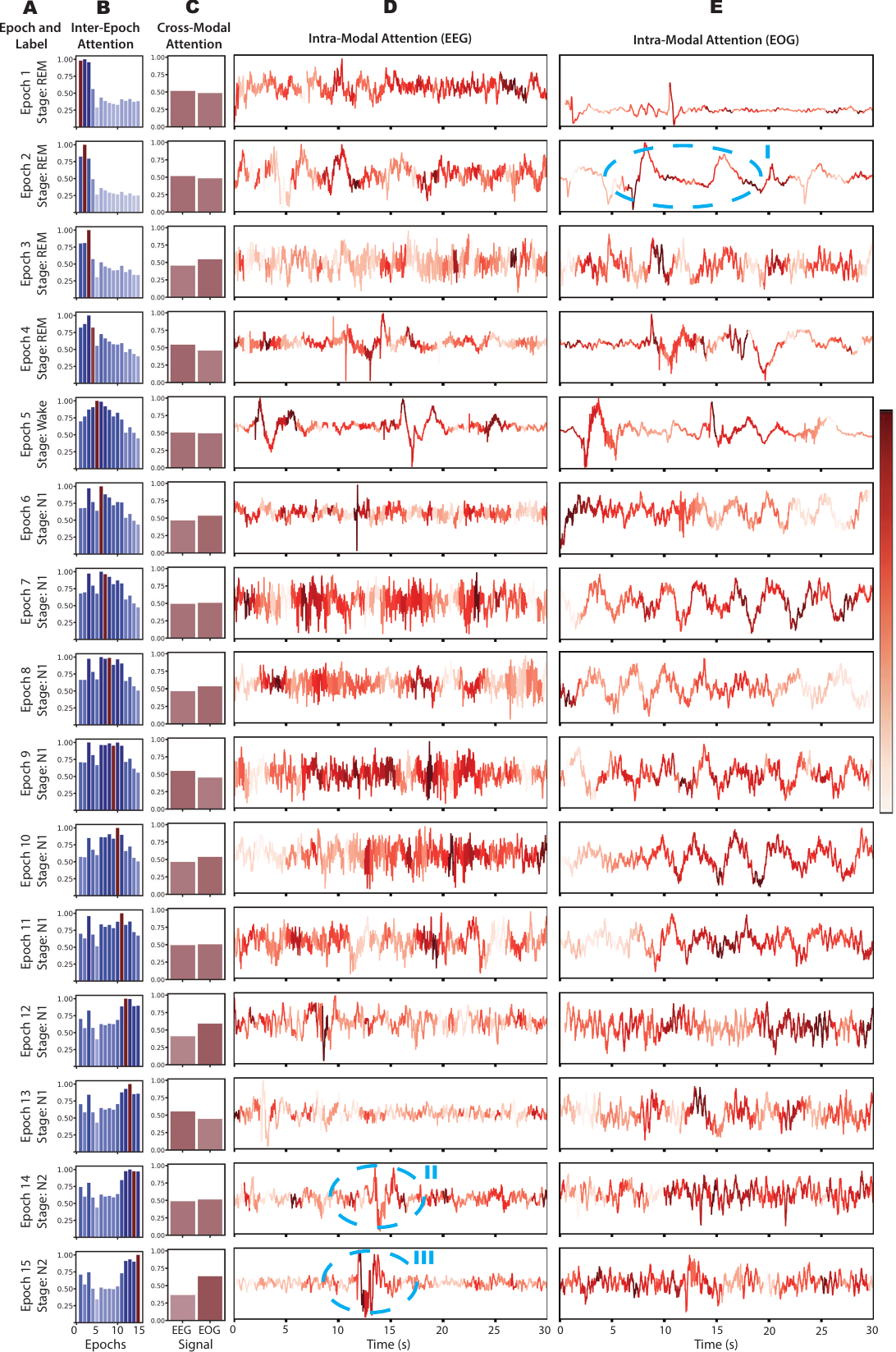}
\vspace{-0.5cm}
\caption{Interpretation results by our sequence cross-modal transformer-15 for a sequence of PSG recordings of subject/recording SC4381 captured between $38820$-$39270$s. Each row in the figure corresponds to a $30$s PSG epoch in the sequence. The column A indicates the corresponding sleep stage of the epoch, while the subsequent columns represent B: inter-epoch attentions, C: cross-modal attentions, D: intra-modal attentions in EEG, and E: intra-modal attentions in EOG, respectively. In inter-epoch attention plots, red bar indicates the normalized self-attention of the current epoch. Additionally, we have highlighted some interesting signal patterns identified by our method using light blue dashed circles with numbers.}
\vspace{-0.5cm}
\label{fig:interpret_15epochs}
\end{figure}

\subsubsection{Comparison Between Epoch and Sequence Cross-Modal Transformers} It can be clearly observed that the sequence cross-modal transformer significantly outperforms epoch cross-modal transformer, because of the inclusion of inter-epoch attention block and predicting sleep stages of multiple epochs simultaneously. Table ~\ref{table: ablation_study_2} clearly indicates the effectiveness of sequential modeling in sleep staging, as it consistently improves the performance over epoch models under various settings. Due to smaller model print, epoch cross-modal transformer is suitable for resource constrained environments, where the algorithm can be implemented on edge devices. 


\subsubsection{Ablation on representation learning}
{The importance of analyzing the representation learning component in our method is important for understanding the construction of classification models in this domain. To gain insights into this aspect, we conducted an ablation study comparing the performance of 1D-CNN, short-time Fourier transform (STFT), and multi-scale 1D-CNNs on the SHHS dataset using an epoch cross-modal transformer. The STFT was used directly with our model without an additional linear input layer. To further analyze the effect of cross-modal attention used in our method, we conducted an additional experiment where cross-modal attention was removed from the final setup with the multi-scale 1D-CNN. The results of these experiments are presented in Table~\ref{table: ablation_study_CNN}. Further analysis of the hyperparameters in our method is provided in the supplementary materials.}
\begin{table}[t]
\centering
\caption{{\bf{Ablation study on 1D-CNN, multiscale-CNN, and STFT as representation learning component on SHHS dataset using epoch cross-modal transformer.}}}

\resizebox{0.98\columnwidth}{!}{%
\begin{tabular}{l|c|c|c|c|c|c|c}
\hline  

\multirow{2}{*}{\textbf{Method}} & \multirow{2}{*}{\textbf{ ACC}} & \multirow{2}{*}{\textbf{$\kappa$}} & \multicolumn{5}{c}{\textbf{ {  Per-class Performance}}} \\ \cline{4-8}
& &&  { W}&  { N1} &  { N2} &  { N3} &  { REM} \\
\hline

\hline

STFT & 74.4 & 0.641 &  86.8 & 8.64 & 76.8 & 70.4 & 59.8 \\

1-D CNN & 78.6 & 0.703 & 89.3 & 22.1 & 79.9 & 75.1  & 70.3 \\

MS-CNN & \textcolor{red}{79.3} & \textcolor{red}{0.712} & \textcolor{red}{89.7} & 20.9 & \textcolor{red}{80.8} & \textcolor{red}{75.7} & \textcolor{red}{71.0} \\
\hline
MS-CNN  & \multirow{2}{*}{78.6} & \multirow{2}{*}{0.702} & \multirow{2}{*}{89.2} & \multirow{2}{*}{\textcolor{red}{22.9}} & \multirow{2}{*}{80.0} & \multirow{2}{*}{75.6} & \multirow{2}{*}{69.8} \\
without CMA &&&&&&&\\
\hline

\end{tabular}}
\begin{flushleft}
 {\scriptsize{- STFT refers to short-time Fourier transform.\\
  - CMA refers to cross-modal attention and MS-CNN refers to multi-scale 1D-CNN.\\
  - In MS-CNN without CMA, we removed CMA from the epoch cross-modal transformer to study its impact on the SHHS dataset. 
  }}
\end{flushleft}

\label{table: ablation_study_CNN}
\vspace{-0.8cm}
\end{table}

{These results clearly indicate that the representations learned by multi-scale CNNs outperform those learned by STFT and 1D-CNN in terms of performance metrics. Furthermore, there is a noticeable drop in performance when cross-modal attention is removed from the model. The improvement of performance achieved by cross-modal attention in both datasets confirms its effectiveness in sleep stage classification tasks.}

\vspace{-0.35cm}
\subsection{Interpreting the  Results}

Along with competitive sleep stage classification performance, the major contribution of cross-modal transformers are their interpretability. As mentioned in section~\ref{subsec:interpret}, we leverage the attention mechanisms to interpret the results as intra-modal, cross-modal and inter-epoch relationships illustrated in Fig.~\ref{fig:interpret_15epochs}. Here, we interpret the predictions of the sequence cross-modal transformer for a sequence of PSG epochs. Specifically, Fig.~\ref{fig:interpret_15epochs} depicts the interpretation results for the sequence cross-modal transformer-15, which is designed to predict 15 epochs in a sequence. 


The interpretation results in Fig.~\ref{fig:interpret_15epochs} focus on a sequence of PSG epochs that include transitions from REM, Wake, N1, and N2 stages.  The inter-epoch attention graphs in Fig.~\ref{fig:interpret_15epochs}B highlight some significant relationships between these epochs. To illustrate, it is evident that the initial four epochs, corresponding to the REM stage, exhibit a notably stronger interconnection compared to the other epochs. This is depicted by the higher inter-epoch attention scores assigned to these first four bars in the bar-plot for the initial four epochs. Similar observation can be seen for N1 stages in from epoch 6 to 13. Moreover, the prediction of the wake stage is related to its preceding epochs (REM) and following epoch (N1) as shown in the bar-plot of epoch 5. Cross-modal attention graphs in Fig.~\ref{fig:interpret_15epochs}C interprets the impact of each modality on the prediction. Finally, the intra-modal attention maps interprets, which time segments in the PSG signal were given more importance by the model for the predictions. In this context, we present a color plot superimposed onto the signal plot, depicting the intra-epoch attention scores associated with various signal regions.  The darker regions highlighted within the signals correspond to higher intra-epoch attention scores, illustrating their importance in the final prediction. Some of such patterns are marked in Fig.~\ref{fig:interpret_15epochs} using light blue dashed line circles with a number. Circle I in Fig.~\ref{fig:interpret_15epochs}E highlights the eye movements which are related to REM sleep. Also, circles II and III in Fig.~\ref{fig:interpret_15epochs}D highlights the K-complex patterns in the N2 sleep stages. By offering interpretablility, our method enables clinicians to further validate the predictions, thereby enhancing reliability.

\vspace{-0.3cm}
\subsection{Limitations}

Despite the potential benefits of transformer-based algorithms for sleep staging, they have several limitations. One major limitation is their high data requirements, often needing large amounts of labeled data for training, which can be challenging with smaller or imbalanced datasets. Strategies like supervised or self-supervised pre-training  \cite{Bert, SIMCLR} can help mitigate this issue. Our method combines transformers with CNNs, reducing data requirements. Another limitation is the computational complexity of transformer-based algorithms, which can demand significant resources for training and inference.



\vspace{-0.3cm}
\section{Conclusion}
In this paper, we present an interpretable, transformer-based deep learning method named \emph{cross-modal transformers}, for automatic sleep stage classification. Our sequence cross-modal transformer achieved performance on par with the state-of-the-art, with reduction in model print and faster training. In addition to accurate sleep staging, our major contribution is to eliminate the black-box behavior of deep learning by leveraging attention mechanisms to interpret the results. We believe that developing interpretable deep learning methods is the most feasible way forward to use artificial intelligence for clinical applications. In future work, the proposed method can be further improved by employing training strategies in deep learning such as transfer learning and self-supervised pre-training.





\vspace{-0.3cm}
\section*{Acknowledgment}
The authors thank Dr. Ranga Rodrigo and the National Research Council of Sri Lanka for providing computational resources.

\vspace{-0.3cm}
\bibliographystyle{IEEEtran}
\IEEEtriggeratref{40}
\bibliography{main}
\newpage
\appendix
\beginsupplement

\section*{Experiments}

\subsection{Data Preprocessing}
\label{datapreprocessing}
\textbf{SleepEDF: }Among the extracted epochs, $69$\% comprised of wake stage. This is due to the existence of long wake stages at the beginning and end of each dataset. Thus, to reduce the redundancy of wake stage, we only considered $30$ minutes of wake stages before and after the sleep periods. Hence, in the resultant dataset, number of epochs are reduced to $196,350$. For our experiments, we considered EEG from the Fpz-Cz channel and horizontal EOG data. All these data were initially segmented to a sequence of $30$~s epochs and corresponding labels were assigned based on the provided annotations. The EEG data was bandpass filtered between $0.2$~Hz to $40$~Hz using a zero-phase finite impulse response filter with a Hamming window. Then, the signals were normalized such that each signal had a zero mean and unit variance. We note that these preprocessing techniques are similar to those employed in state-of-the-art methods~\cite{Trans1,xsleepnet}.

\textbf{SHHS:}  We used EEG signals from the C4-A1 channel and horizontal EOG (L-R) signals from the SHHS PSG montage. Similar to the sleepEDF dataset, we segmented the data into 30-second epochs, with labels assigned based on the provided annotations. Both the EEG and EOG signals were normalized using the 0.95 quantile value, ensuring that the normalization process is less affected by outliers (i.e., transients).

\vspace{-0.3cm}
\subsection{Training Setup}

 We defined two experimental setups based on the classification schemes (one-to-one and many-to-many) utilizing our proposed epoch and sequence cross-modal transformers. Both models were trained using the Adam optimizer~\cite{kingma2014} with the learning rate ($lr$), $\beta_1$ and $\beta_2$ set to, $10^{-3}$, $0.9$ and $0.999$, respectively. A weight decay of $0.0001$ was applied to avoid overfitting. The batch size was experimentally chosen to $32$ for sleepEDF and $256$ for SHHS. We used weighted categorical cross entropy as the loss function ($\mathcal{L}$) for $5$-class classification and the weights for the classes were set to $\{1, 2, 1, 2, 2\}$, to handle the data imbalance. For the sequence cross-modal transformer, we empirically chose the number of epochs in an input sequence as $5$, $10$, $15$ and $21$. For training and evaluation of both proposed methods, we divided the SleepEDF-78 dataset into five subject-independent groups, i.e., the dataset was randomly split into five different groups subject wise. Then, $5$-fold cross validation was used to evaluate the performance of the model, and to tune both hyper-parameters and the model architecture. For SHHS, we randomly split the subjects in the dataset into 70\% for training and 30\% for testing, with 100 subjects from the training set separated as a validation set. The model was implemented in the Pytorch environment and trained using a Nvidia Quadro RTX $5000$ graphics card with $16$ GB memory.



\vspace{-0.3cm}
\subsection{Evaluation Metrics}
Accuracy (ACC), Cohen's kappa coefficient ($\kappa$), macro averaged F1-score (MF1), sensitivity (Sens.), specificity (Spec.) and macro averaged G-mean (MGm) are the metrics used to evaluate our method. Macro averaged F1-score is calculated as the average of F1-scores of all 5 classes. MF1 and MGm are considered because they are suitable metrics to evaluated the performance on imbalanced datasets. 
Additionally, we report F1-scores for each sleep stages to further evaluate the model's performance. Given the true positives (TP), true negative (TN), false positive (FP) and false negative (FN), MF1, Sens., Spec. and MGm can be calculated as follows,


\small{
$$ MF1 = 2\frac{Precision \times Sensitivity}{Precision + Sensitivity}, $$ 
$$ MGm = \sqrt{Specificity \times Sensitivity}.$$\\
\vspace{-2em}
where \\
\vspace{-1em}
$$Precision = \frac{TP}{TP+FP},$$\\
\vspace{-1em}
$$ Sens. = \frac{TP} {TP+FN}, \hspace{2em} Spec. = \frac{TN} {TN + FP}$$\\
\vspace{-1em}}

\section*{Results and Discussion}

\subsection{Hyper-parameter Tuning}
\begin{table}[!t]
\centering
\caption{{\bf Ablation study on different model variations of epoch cross-modal transformer by varying embedding dimension ($E$) and dimension of the hidden layer in feed forward network ($D_{ff}$).}}

\vspace{-.5em}
\begin{tabular}{c|c|c|c}
\hline  

\textbf{Embedding}&\textbf{Dim feed} &  \multirow{2}{*}{\textbf{ACC}} &  \multirow{2}{*}{\textbf{$\kappa$}}\\
\textbf{Dim ($E$)}&\textbf{Forward ($D_{ff}$)}&&\\
\hline
 
$64$ & \multirow{3}{*}{$256$} & $80.66$ & $0.735$\\
$128$ &  &$80.47$ & $0.733$\\
$256$ &  & $80.63$ & $0.734$\\
\hline
$64$ & \multirow{3}{*}{$512$} & $80.39$ & $0.732$\\
$128$ &  & \textcolor{red}{80.75} & \textcolor{red}{0.736}\\
$256$ &  &$80.52$ & $0.732$\\
\hline
$64$ & \multirow{3}{*}{$1024$} & $80.49$ & $0.733$\\
$128$ &  & $80.63$ & $0.735$\\
$256$ &  &$80.44$ & $0.732$\\

\hline
\end{tabular}
\vspace{-0.1cm}
\label{table: ablation_study}
\end{table}
\begin{table}[!t]
\centering
\caption{{\bf Ablation study on different model variations of epoch cross-modal transformer by varying attention blocks.}}

\begin{tabular}{c|c|c|c|c}
\hline  

\textbf{Attention}&\textbf{Embedding}&\textbf{Dim feed} &  \multirow{2}{*}{\textbf{ACC}} &  \multirow{2}{*}{\textbf{$\kappa$}}\\
\textbf{Blocks}&\textbf{Dim ($E$)}&\textbf{Forward ($D_{ff}$)}&&\\
\hline

1&\multirow{4}{*}{$128$} & \multirow{4}{*}{$512$} &$80.8$  & $0.736$ \\
2 & &  & $80.5$& $0.733$\\
 3&&  &$80.5$ &$0.732$ \\
  4&&  &$80.3$& $0.730$\\
\hline

\end{tabular}
\vspace{-0.2cm}
\label{table: ablation_study_attention}
\end{table}
Ablation study was conducted to identify the most suitable model variation to achieve the optimal performance with minimal model print. The study was conducted on epoch cross-modal transformers, where the embedding dimension ($E$ $\in$ $\{64, 128, 256\}$) and the number of neurons in the hidden layer of the -forward networks ($D_{ff}$ $\in$ $\{256, 512, 1024\}$) were varied.  The epoch cross-modal transformers are used to conduct the study efficiently and, the same parameters were adapted to sequence cross-modal transformers. Tuning the hyper-parameters on epoch cross-modal transformers and then scaling it towards sequence cross-modal transformers is more efficient compared to tuning the hyper-parameters on the level of sequence cross-modal transformers. The results of the study is provided in Table \ref{table: ablation_study}, where the model variation with $E = 128$ and $D_{ff} = 512$ achieved the best performance. A study was conducted using epoch cross-modal transformer with $E = 128$ and $D_{ff} = 512$ to find the most suitable number of attention blocks in the intra-modal and cross-modal blocks. As per Table \ref{table: ablation_study_attention}, there is no significant improvement in performance with the increase of the attention blocks. Based on the hyper-parameter tuning study, $E = 128$, $D_{ff} = 512$, and one attention block per intra-modal and cross-modal blocks were used in the sequence cross-modal transformer to achieve good performance with a small model footprint.

\end{document}